\documentclass{article} 

\usepackage{iclr2018_conference,times}
\usepackage{url}
\usepackage{amssymb}
\usepackage{subcaption}
\usepackage{graphicx}
\usepackage{float}
\usepackage{bbm}
\usepackage[noend]{algpseudocode}
\usepackage{amsmath}
\usepackage{amsthm}
\usepackage{enumitem}
\usepackage[]{algorithm2e}
\usepackage{subcaption}
\newtheorem{prop}{Proposition}
\usepackage[usestackEOL]{stackengine}
\usepackage{titlesec}
\usepackage{float}
\usepackage{amsmath}
\usepackage[colorlinks=true,linkcolor=blue]{hyperref}%
\usepackage{hyperref}       
\usepackage{url}            
\usepackage{booktabs}       
\usepackage{amsfonts}       
\usepackage{nicefrac}       
\usepackage{microtype}      
\usepackage{enumitem}
\usepackage{graphicx}
\usepackage{float}
\usepackage{standalone}
\usepackage{lipsum}
\usepackage{adjustbox}
\usepackage{verbatim}
\usepackage{mathabx}
\usepackage{booktabs} 
\usepackage{latexsym,eucal,color}
\usepackage{siunitx}

\title{Sparse Multi-Family Deep Scattering Network}

\iclrfinalcopy
\author{Romain Cosentino, Randall Balestriero  \\
Department of Electrical and Computer Engineering\\
Rice University\\
 6100 Main St, Houston, TX 77005 \\
}

\begin{document}
\maketitle

\begin{abstract}
In this work, we propose the Sparse Multi-Family Deep Scattering Network (SMF-DSN), a novel architecture exploiting the interpretability of the Deep Scattering Network (DSN) and improving its expressive power. The DSN extracts salient and interpretable features in signals by cascading wavelet transforms, complex modulus and extract the representation of the data via a translation-invariant operator. First, leveraging the development of highly specialized wavelet filters over the last decades, we propose a multi-family approach to DSN. In particular, we propose to cross multiple wavelet transforms at each layer of the network, thus increasing the feature diversity and removing the need for an expert to select the appropriate filter. Secondly, we develop an optimal thresholding strategy adequate for the DSN that regularizes the network and controls possible instabilities induced by the signals, such as non-stationary noise. Our systematic and principled solution sparsifies the network's latent representation by acting as a local mask distinguishing between activity and noise. The SMF-DSN enhances the DSN by $(i)$ increasing the diversity of the scattering coefficients and $(ii)$ improves its robustness with respect to non-stationary noise.
\end{abstract}

\section{Introduction}
\label{Intro}
Modern machine learning focuses on developing algorithms to tackle natural machine perception tasks such as speech recognition, computer vision, recommendation, among others \citep{deng2013new,lecun2015deep,noda2015audio,strub2016hybrid, mu2018survey}. 
Historically, some of the proposed models were based on well-justified mathematical tools from signal processing such as Fourier analysis
\citep{bou2000comparative,schluter2001using,bicego2006use}
. However, such theory-guided approaches have become almost obsolete with the growth of computational power and the advent of high-capacity models. Over the past decade, the standard solution evolved around deep neural networks (DNNs) \citep{goodfellow2016deep}. 
While providing state-of-the-art performance on many benchmarks, at least two pernicious problems still plague DNNs:
First, the absence of stability in the DNN's input-output mapping has famously led to adversarial attacks where small perturbations of the input lead to dramatically different outputs \citep{goodfellow2014explaining,kurakin2016adversarial,athalye2018synthesizing}. In addition, this lack of control manifests in the detection thresholds (i.e: ReLU bias) of DNNs, rendering them prone to instabilities when their inputs exhibit non-stationary noise and discontinuities. 
To alleviate these issues, we propose the use of the Deep Scattering Network (DSN) \citep{bruna2013invariant} which provides an efficient hand-crafted representation of the data that is theoretically guided, interpretable, and computationally efficient as it does not require any learning \citep{seydoux2020clustering}.

One of the main differences between DNNs and the DSN is that DSN makes use of hand-crafted wavelet filters. The same way that the convergence toward optimal filter in DNN matters, the selection of which wavelet filter is selected will impact its performance. A large number of wavelet filters have been designed, and each of them yields specific time-frequency properties \citep{torrence1998practical}. In particular, their time and frequency localization governed by the Heisenberg principle induces their suitability with respect to a given application. For instance, the detection of time transient can be eased by selecting a filter highly localized in time and can be tedious with a filter having a small frequency support. This selection usually necessitates the intervention of an expert on the signal at hand or one should consider their learnability \citep{gilles2013empirical,zeghidour2018learning,balestriero2018spline,Cosentino2020LearnableGT}. In this work, we propose to avoid the computational burden induced by an adaptive or learnable filter and alleviate the requirement for an expert by considering a multi-family approach. Different wavelet families are presently used to extracting all the signal's salient features at hand.

We develop a network topology crossing the different wavelet transforms to guarantee that at least one of the network's path can adequately capture all the important events in the signal. In fact, the importance of a wavelet family can be tied to a specific layer. The concatenation of different family appears to be sub-optimal as the patterns to be encoded at each layer vary significantly due to the composition of transformations inherent to the DSN. This cross multi-family approach yields the Multi-Family Scattering Network (MF-DSN).

Also, while such a DSN's filtering operation is not learned, which guarantees its stability and robustness, one can consider improving its representative power by proposing an adaptive and theoretically justified nonlinearity. While traditional DNNs benefit from signal-adapted nonlinearities such as the ReLU, the DSN does not provide any adaptive mechanisms with respect to the input signal. We propose to exploit the recent development of a universal frame thresholding algorithm to equip the DSN with an adaptive nonlinear operator removing nuisances in the representation at each layer of the network \citep{thresh}. Considering both the multi-family framework and the thresholding algorithm, we build the Sparse Multi-Family Scattering Network (SMF-DSN).


The paper's organization is as follows: In Sec.~\ref{sect2}, we relate the various developments around the DSN as well as thresholding frameworks. Then in Sec.~\ref{contrib}, we elaborate on the contributions of the present paper. Then, in Sec.~\ref{DeepCroisee} we perform the construction of both the MF-DSN architecture and the thresholding algorithm. In Sec.~\ref{EventCharacterization} we show that such a network can be used to characterize different audio events on the Freefield1010\footnote{http://machine-listening.eecs.qmul.ac.uk/bird-audio-detection-challenge/}  audio scenes dataset. Finally, we evaluate our architecture on a bird detection task in Sec.~\ref{validation}.

\subsection{Related Work}
\label{sect2}

We extend the DSN, first developed in \citet{mallat2012group} and successfully applied in \citet{bruna2011classification,anden2011multiscale}. It consists in a cascade of linear and nonlinear operators applied on the input signal. The linear transformation is a wavelet transform, and the nonlinear transformation is a complex modulus. At each layer, the convolution of the scalogram with a scaling function leads to the so-called scattering coefficients.
This network is stable (Lipschitz-continuous) and suitable for machine learning tasks as it removes spatio-temporal nuisances by building space/time-invariant features.
The translation invariance property is provided by the scaling function that acts as an averaging operator on each layer of the transform leading to an exponential decay of the scattering coefficients \citep{waldspurger2017exponential}. Since the wavelet transform increases the number dimension of the signal, the complex modulus is used as its contractive property reduces the variance of the projected space \citep{mallat2016understanding} as well as to steer the signal information toward the low frequency content so that the scattering coefficients are capturing the signal's information. 
Among the extensions of this architecture, the closest to this work are: the Joint DSN \citep{anden2015joint} and the time-chroma-frequency DSN\citep{lostanlen2016wavelet}. In both work,
they introduced an extra parameterization of the wavelets coefficients in the second layer of the network to capture frequency correlations allowing the scattering coefficient to represent the transient structure of harmonic sounds.

Thresholding in the wavelet domain remains a powerful approach for signal denoising as it exploits the edge-detector property of wavelets, providing a sparse representation of the input signal in the time-frequency plane. This property is characterized for each wavelet by its vanishing moments expressing the wavelet's orthogonality to a given smoothness order in the input signal. 
We base our approach on the theories relating the thresholding of signal in the wavelet basis and evaluating the best basis. 
Both are realized via a risk evaluation that arose from different perspectives:  statistical signal processing \citep{donoho1994ideal,donoho1995wavelet,krim1999denoising}, information theory \citep{coifman1992entropy,IQR,moi}, and signal processing \citep{mallat1993matching,Wavtour}. In this work we will opt for a universal frame thresholding technique developed in \cite{thresh} particularly suited for the DSN.

\subsection{Contributions}
\label{contrib}
As opposed to the chroma-time-frequency DSN, using one wavelet family filter bank but deriving tensor product of the latter, we propose to use multiple wavelet families having complementary properties (described in Appendix~\ref{family}) within a unified network yielding cross-connections. 
Such a multi-family approach provides higher dimensional and uncorrelated features, reducing the need for an expert to hand-choose the appropriate wavelet filters.

Therefore our architecture, the Multi-Family Deep Scattering Network (MF-DSN), leverages the simultaneous decomposition of complementary filter-banks and their crossed decomposition. 
Then, endowing this architecture with an overcomplete thresholding operator, we build the SMF-DSN, providing a novel nonlinearity based on each wavelet dictionary's reconstruction risk. The thresholding method, based on empirical risk minimization, will bring several advantages:
\begin{enumerate}
\item It enables us to ensure and control the stability of the input-output mapping via thresholding the wavelet coefficients.
\item The model has sparse latent representations that ease the learning of decision boundaries and increases generalization performance.
\item The risk associated with each wavelet family provides a characterization of the time-frequency components of the analyzed signal, that, when combined with scattering features, enhances the linearization capacity of DSN.
\end{enumerate}
As opposed to ReLU-based nonlinearities that impose sparsity by thresholding coefficients based on a fixed learned scalar threshold, we propose an \textit{input-dependant locally adaptive} thresholding method. 
Therefore, our contribution leading to the Sparse Multi-Family Deep Scattering Network is twofold: 
\begin{itemize}
    \item We propose a natural extension of the DSN, allowing the use of multiple wavelet families and their crossed representations at each layer of the network.
    \item We derive an optimal frame thresholding technique in which the empirical risk minimization leads to an analytical solution endowing the DSN with sparse latent representations.
\end{itemize}

\section{Sparse Multi-Family Deep Scattering Network}
\label{DeepCroisee}

The Multi-Family Deep Scattering Network is a tree architecture ($2$ layers of such model are shown in Fig. \ref{fig:DCS}) based on the Deep Scattering Network (DSN). The first layer of a scattering transform corresponding to the standard scalogram is now replaced with a  $3$-dimensional tensor by adding the wavelet family dimension. Hence, it can be seen as a stacked version of multiple scalograms, one per wavelet family. The second layer of the MF-DSN brings inter and intra wavelet family decompositions. Each wavelet family of the second layer will be applied on all the first layer scalograms; the same process is successively applied for building a deeper model.

\subsection{Multi-Family Deep Scattering Network}
\label{network}

We first proceed by describing the formalism of the MF-DSN. Note that details on wavelets and filter-bank designs are provided in Appendix~\ref{appendixA}.

We denote by 
\begin{equation}
\Psi^{(1)}=\{\psi^{(1,b)},b=1,\dots,B^{(1)}\},
\end{equation} 
the collection of $B^{(1)}$ wavelet families for the first layer. We also denote by,
\begin{equation}
\lambda^{(1)} = \left \{ \lambda^{(1)}_j, j= 1,...,J^{(1)} \times Q^{(1)} \right \},\;\; \lambda_j^{(1)}=2^{(j-1)/Q^{(1)}},
\end{equation}
the resolution coefficients for this first layer where $J^{(1)}$ represents the number of octave and $Q^{(1)}$ the quality coefficients a.k.a the number of wavelets per octave.

Based on those configuration coefficients, the filter-banks can be derived by scaling of the mother wavelets with respect to the resolution coefficients. We thus denote the filter-bank creation operator $\mathcal{W}$ and defines the first layer filter-bank as
\begin{align}
\mathcal{W}[\psi^{(1,b)},\lambda^{(1)}]=\begin{pmatrix}
\psi^{(1,b)}_{\lambda_1^{(1)}}\\
\dots \\
\psi^{(1,b)}_{\lambda_{J^{(1)} \times Q^{(1)}}^{(1)}}\\
\end{pmatrix},\;\;\psi^{(1,b)}_{\lambda^{(1)}_j}(t)=\frac{1}{\sqrt{\lambda^{(1)}_j}}\psi^{(1,b)}\left(\frac{t}{\lambda^{(1)}_j}\right).
\end{align}
That is, a filter-bank $\mathcal{W}$ depends on a mother wavelet $\psi^{(1,b)}$, i.e., the $b^{th}$ mother wavelet for the first layer, and $\lambda^{(1)}$ the set of first layer set of resolution coefficients.

\begin{figure}[t]
    \centering
    \includegraphics[width=\linewidth]{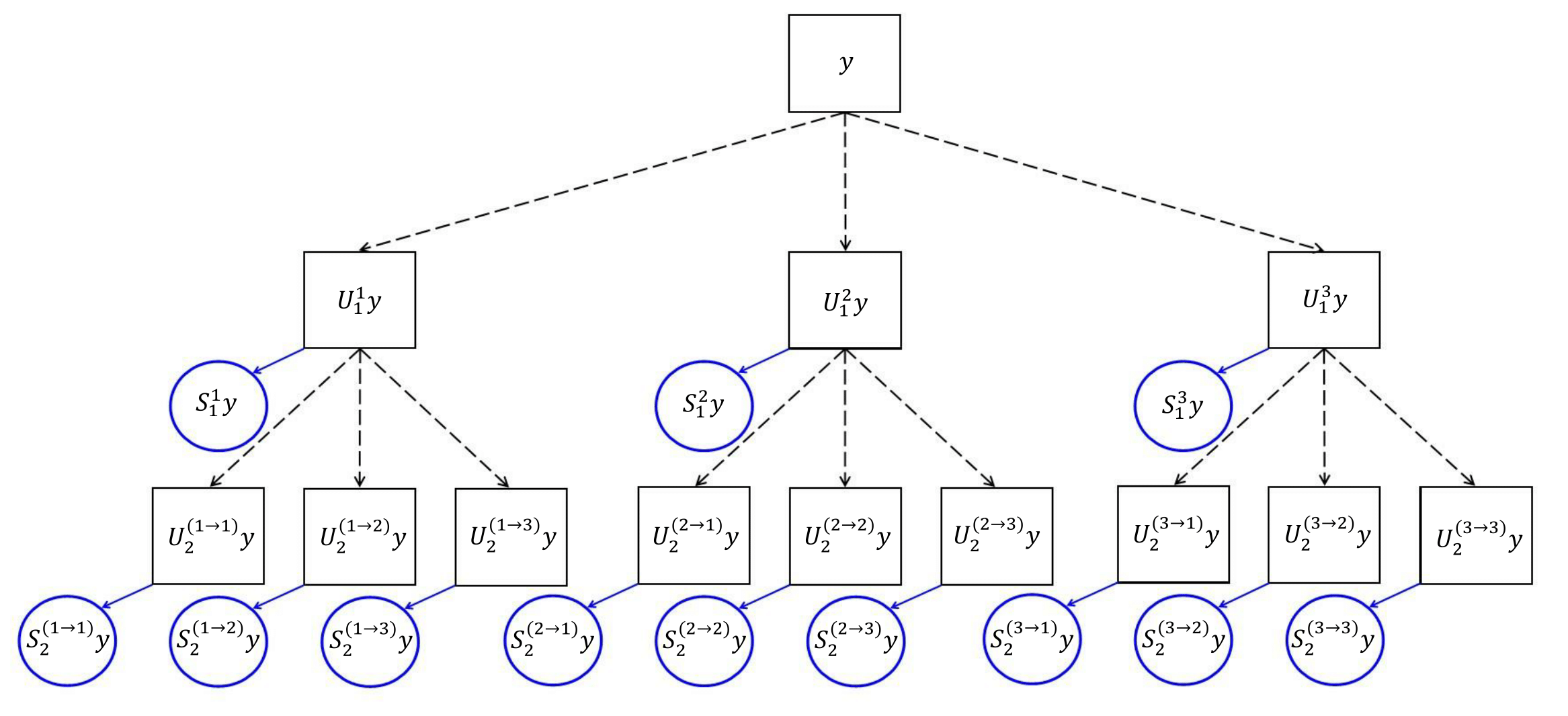}
    \caption{Multi-Family Deep Scattering Network Architecture for an input signal $y$ - Special case of $3$ wavelet families.}
    \label{fig:DCS}
\end{figure}

To avoid redundant notation, we denote this filter-bank as $\mathcal{W}^{(1,b)}$ with implicit parameters $\psi^{(1,b)}$ and $\lambda^{(1)}$.

We now developed the needed tools to define the filter layer of the MF-DSN explicitly. We denote by $U^{(1)}$ the output of this first layer, and as previously mentioned, it consists of a $3D$ tensor of shape $(B^{(1)},J^{(1)}Q^{(1)},N)$ with $N$ the length of the input signal denoted as $y$ and is defined by
\begin{align}
U^{(1)}[y](b,.,.)=|y \star \mathcal{W}^{(1,b)}|, b=1,\dots,B^{(1)},
\end{align}
where $|.|$ operator corresponds to an element-wise complex modulus application and the convolution is defined as
\begin{align}
y \star \mathcal{W}^{(1,b)}= \begin{pmatrix}
y \star \mathcal{W}^{(1,b)}(1,.) \\
\dots \\
y \star \mathcal{W}^{(1,b)}(J^{(1)}Q^{(1)},.) 
\end{pmatrix}.
\end{align}
For each wavelet family $b \in \left \{1,\dots, B^{(1)} \right \}$ one compute a time-frequency representation $U^{(1)}[y](b,.,.)$ where the first dimension corresponds to the number of filters in the filter bank, and the second is the time dimension that depends on the size of the input signal.

The second layer contains cross terms, that is, the time-frequency transformation for all wavelet families of the second layer applied to all the time-frequency representations of the first layer.
We denote the second layer representation as $U^{(2)}[U^{(1)}[y]]$. This object is a $5D$ tensor introduced $2$ extra dimension on the previous tensor shape, i.e., $U^{(1)}[y]$ and is defined as
\begin{align}
U^{(2)}[U^{(1)}[y]](b_2,j_2,b_1,.,.)= | U^{(1)}(b_1,.,.) \star \mathcal{W}^{(\ell,b_2)}(j_2)|,
\end{align}
where $b_2=1,\dots,B^{(2)}, j_2 \in \left \{1,\dots, J^{(2)}Q^{(2)} \right \}$. Thus, $U^{(2)}[U^{(1)}[y]](b_2,j_2,b_1,.,.)$ denotes the filtration of time-frequency representation $U^{(1)}[y](b,.,.)$, with respect to the filter $\psi^{(j_2,b2)}$. That is, the last two dimensions of this operator correspond to $J^{(1)}Q^{(1)}$, i.e., the resolution of the first layer time-frequency representation, and $N$ the number of sampled contained in the input signal. 

Note that the cross representation are all the terms in $U^{(2)}[U^{(1)}[y]](b_2,j_2,b_1,.,.)$ where $b_2 \neq b_1$, that is, where the filter at the second layer belongs to the family $b_2$ and the filter of the first layer belongs to the family $b_1$.

For notations clarity we denote this representation as $U^{(2)}[U^{(1)}[y]](b_2,j_2,b_1,.,.):=U^{b_1 \rightarrow b_2}_{j_2}[y]$, that is, the path starting at the scalogram induced by the wavelet family $b_1$ and ending at the $j_2^{th}$ filter belonging to family $b_2$ in the second layer. This notation is easily extended to generalation to the any path going from layer $1$ to layer $\ell$ as $U^{b_1 \rightarrow \dots \rightarrow b_\ell}_{j_2,\dots,j_\ell}[y]$. We however limit ourselves in practice to $2$ layers as usually done with the standard scattering networks.

Given these representations, the scattering coefficients are defined as follows:
\begin{equation}
\begin{matrix}
S^{b_1 \rightarrow \dots \rightarrow b_\ell}_{j_2,\dots,j_\ell}[y] = U^{b_1 \rightarrow \dots \rightarrow b_\ell}_{j_2,\dots,j_\ell}[y] \star \phi,
\end{matrix}
\end{equation}
where $\phi$ is a scaling function (low-pass filter). The application of a low frequency band-pass filter allows for symmetries invariances, inversely proportional to the cut-off frequency of $\psi$. In particular it yields the translation invariance property of the DSN. The schematic representation of a Multi-Family Deep Scattering Network for two layers is shown in Fig.~\ref{fig:DCS}.

Let's now tackle the problem of thresholding an overcomplete basis, cases where the quality factor, $Q$, is greater than $1$, which are in practice needed to design a filter-bank providing enough frequency precision to the signal representation. 
\subsection{Sparsity and Winner-take-all via Risk Minimization}
\label{denoise}
Sparsity in the latent representation of network models have been praised many times \citep{narang2017exploring,liu2015sparse,thom2013sparse}. It represents the fitness of a model's internal parameters with only a few nonzeros coefficients to perform the task at hand. Furthermore, sparsity is directly related to the Minimum Description Length \citep{Dhillon:2011:MDL:1953048.1953064} guaranteeing increased generalization performances. Besides those concepts, thresholding brings in practice robustness to noise. In particular, as we will demonstrate, even in large scale configuration, non-stationary noise can not be handled entirely by common machine learning approaches on their own.
To do so, we use a recently developed universal thresholding technique for non-orthogonal filter-banks \citep{thresh}, which aim at minimizing the reconstruction error of the thresholded signal in the wavelet domain via an oracle decision. We can derive analytical thresholding formulas based on the input representation and the filter-bank redundancy through this formulation.

\subsubsection{Ideal Risk and Empirical Risk Bound}

As the decomposition is not orthogonal, the first point to be tackle is the difference of the $L^2$ approximation errors in between the original basis and the over-complete wavelet basis as Parseval equality does not hold. Besides, the transpose of the change of basis matrix is not anymore the inverse transform. In \citet{berkner1998correlation} proposed using the Moore pseudo inverse to build the reconstruction dictionary.
In the following, we develop an upper bound to the ideal risk such that we benefit from an explicit equation for the thresholding operator that is adapted to any over-complete transformation.
Let's assume the observed signal, denoted by $y$, is corrupted with white noise such that $y= x +\epsilon$ where $x$ is the signal of interest and $\epsilon \sim \mathcal{N}(0,\sigma^2)$.
We now denote by $W \in \mathbb{C}^{N(JQ+1) \times N}
$ the matrix composed by the the wavelets at each time and scale (i.e: explicit convolution version of $\mathcal{W}$) such that $\forall x \in \mathbb{R}^N$, $Wx$ is the wavelet transform. We denote by $W^{\dagger} \in  \mathbb{C}^{N \times N(JQ+1)}$ the generalized inverse such that  $W^{\dagger}W=I$. 
The estimate of $x$ is given by  $\hat{x}_{W,D}(y) = W^{\dagger}D^{S}Wy$.
\begin{align}
 \mathcal{R}^{\star}(x,W)&=  \min_{\delta} \mathbb{E} \left \| x-\hat{x}_{W,D}(y) \right \|^2 =  \min_{\delta} \mathbb{E} \left \| W^{\dagger} (Wx-D^{S}Wy) \right \|^2 
 \end{align}
Because of the correlation implied by the redundant information contained in the filter banks, the ideal risk is now dependent on all the possible pairs in the frequency axis. However,the independence in time remains. Since this optimization problem does not have an analytic expression, we propose the following upper bound explicitly derived in Appendix \ref{upper-bound}. 
The upper-bound on the optimal risk is denoted by $\mathcal{R}_{up}$ and defined as,
\begin{align}
    \mathcal{R}_{up}(x,W) = \sum_{k =1}^{N(JQ+1)}  \min(\mathcal{R}_{up}^{U}(x),\mathcal{R}_{up}^{S}).
\end{align}
where we denote by $\mathcal{R}_{up}^{U}$ the upper bound error term corresponding to unselected coefficients:
\begin{equation}
    \mathcal{R}_{up}^{U}(x) =\sum_{j =1}^{n*(JQ+1)} \left | \mu_{k}(x)\mu{j}(x) \sum_{t=1}^{n}    \psi^{\dagger}_{t}[k] \psi^{\dagger}_{t}[j]   \right |,
\end{equation}
and by $\mathcal{R}_{up}^{S}$ the upper bound error term corresponding to the selected coefficients:
\begin{equation}
    \mathcal{R}_{up}^{S} = \sigma^{2} \sum_{j=1}^{N(JQ+1)}   \left | \sum_{t=1}^{n} (\psi^{\dagger}_{t}[k] \psi^{\dagger}_{t}[j]) \psi_k^{T} \psi_j  \right |.
\end{equation}
Now, one way to evaluate this upper-bound is to assume an orthogonal basis, and to compare it with the optimal risk in the orthogonal case which leads to the following proposition.
\begin{prop}
Assuming orthogonal filter matrix $W_O$, the upper bound ideal risk coincides with the orthogonal ideal risk:
\begin{equation*}
    \mathcal{R}_{up}(x,W_O) = \mathcal{R}_{O}(x,W_O)
\end{equation*}
\end{prop}
the proof is derived in \ref{upper-bound-comparison}
In order to apply the ideal risk derive, ones needs an oracle decision regarding the signal of interest. In real application, the signal of interest $x$ is unknown. We thus propose the following empirical risk:
\begin{align}
    \tilde{\mathcal{R}}_(y,W) =  \sum_{k =1}^{N(JQ+1)}  \min(\mathcal{R}_{up}^{U}(y),\mathcal{R}_{up}^{S}).
\end{align}
This risk corresponds to the empirical version of the ideal risk where the observed signal $y$ is evaluate in the left part of the minimization function. In order to compare this empirical risk with the ideal version, we propose their comparison the following extreme cases:
\begin{prop}
In the case where $D^S=I$, the empirical risk coincides with the upper bound ideal risk:
\begin{equation*}
   \tilde{\mathcal{R}}(y,W) =   \mathcal{R}_{up}(x,W).
\end{equation*}
\end{prop}
\begin{prop}
In the case where $D^U=I$, the following bound shows the distance between the empirical risk and the upper bound ideal risk:
\begin{equation}
   \tilde{\mathcal{R}}(y,W) \leq \mathcal{R}_{up}(x,W) + C \times \left | \sum_{t=1}^{n}    \psi^{\dagger}_{t}[k] \psi^{\dagger}_{t}[j]   \right |, a.s.
\end{equation}
where, 
\begin{equation*}
    C =  \sum_{k=1}^{N(JQ+1)} \sum_{j=1}^{N(JQ+1)}   \left | \mu_{k}(x) \right | \left \| \psi_j \right \|_1 \sigma \sqrt{\frac{2}{\pi}} +  \left | \mu_{j}(x) \right |\left \| \psi_k \right \|_1 \sigma \sqrt{\frac{2}{\pi}}  + \sigma^2 (1- \frac{2}{\pi}).
\end{equation*}
\end{prop}
Refer to \ref{upper-bound-empirical} for proofs.
As the empirical risk introduces the noise in the left part of the risk expression, this term represents this noise's propagation throughout the decomposition. 
We provided a generic development of the risk minimization process. When applied to a particular path of the scattering network, it is denoted as,
\begin{equation}
R^{b_1 \rightarrow \dots \rightarrow b_\ell}_{j_1,\dots,j_{\ell -1}}[y] =\mathcal{R}[U^{(\ell)}[y](b_\ell,.,b_{\ell-1},j_{\ell-1},\dots,b_1,j_1,.),\mathcal{W}^{(\ell,b_\ell)}],b_\ell =1,\dots,B^{(\ell)},
\end{equation}
with $U^{(\ell)}[y](b_\ell,.,b_{\ell-1},j_{\ell-1},\dots,b_1,j_1,.) \in \mathbb{R}^{J^{(\ell)}Q^{(\ell)}\times n}$ and $\mathcal{R}$ representing the risk minimization operator based on a given representation and the associated filter-bank.

\begin{figure}[t]
    \centering
    \includegraphics[width=\linewidth]{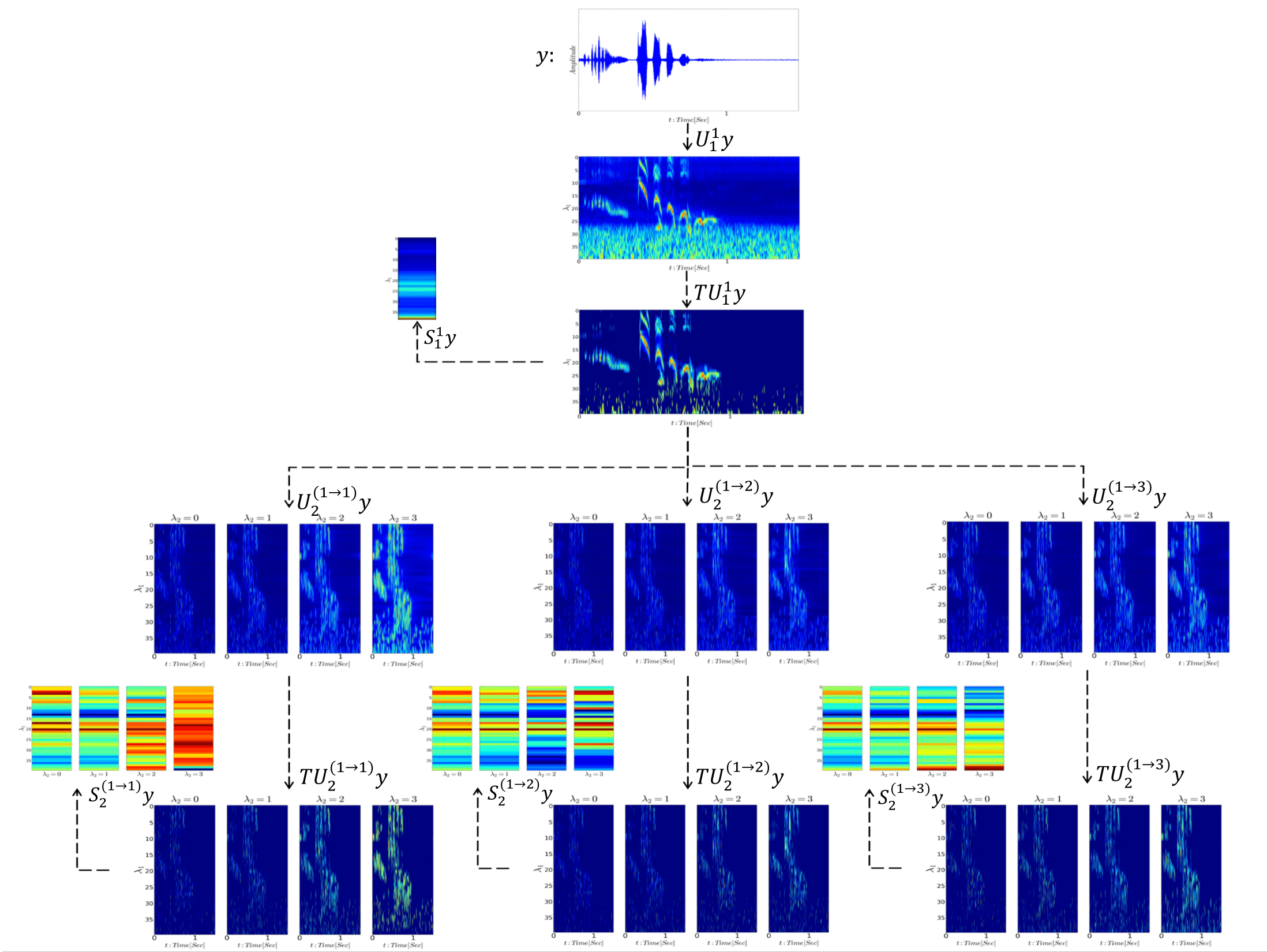}
    \caption{\textbf{Sparse Multi-Family Deep Scattering Network} representation - Paths: for the first layer we display only $1$: Morlet transform. Then for the second layer the paths $1 \rightarrow 1,1 \rightarrow 2,1 \rightarrow 3$ are performed respectively with $1:$ Morlet, $2:$ Paul, $3$: Gammatone - The hyperpameter of each wavelet tranform are J1= 5, Q1=8, J2=4, Q2=1.}
    \label{fig:DS-apply}
\end{figure}

Now, let's denote by $\mathcal{T}$ the tresholding operator minimizing the the empirical risk defined as
\begin{equation}
      \mathcal{T} = \arg \min_{\delta} \tilde{\mathcal{R}}(y,W).
\end{equation}
This operator can then be applied on a specific path of the MF-DSN tree such as
\[ \mathcal{T}[U^{(\ell)}[y](b_\ell,.,b_{\ell-1},j_{\ell-1},\dots,b_1,j_1,.),\mathcal{W}^{(\ell,b_\ell)}],b_\ell =1,\dots,B^{(\ell)}. \]
We provide in Fig. \ref{fig:DS-apply} illustration showing the effect of this thresholding operator at each layer of the MF-DSN.

\subsection{Event Detection}
\label{EventCharacterization}

The empirical evaluation of the risk developed above represents the ability of the induced representation to perform efficient denoising and signal reconstruction. This concept is identical to the one of function fitness when considering the denoised ideal signal $x$ and the thresholded reconstruction. As a result, it is clear that the optimal basis given a signal is the one with minimal empirical risk. We thus propose here simple visualization and discussion on this concept and motivate the need to use the optimal empirical risk as part of the features characterizing the input signal $y$ along all the representations.

\begin{figure}[t]
  \begin{subfigure}[b]{0.5\textwidth}
    \includegraphics[width=\textwidth]{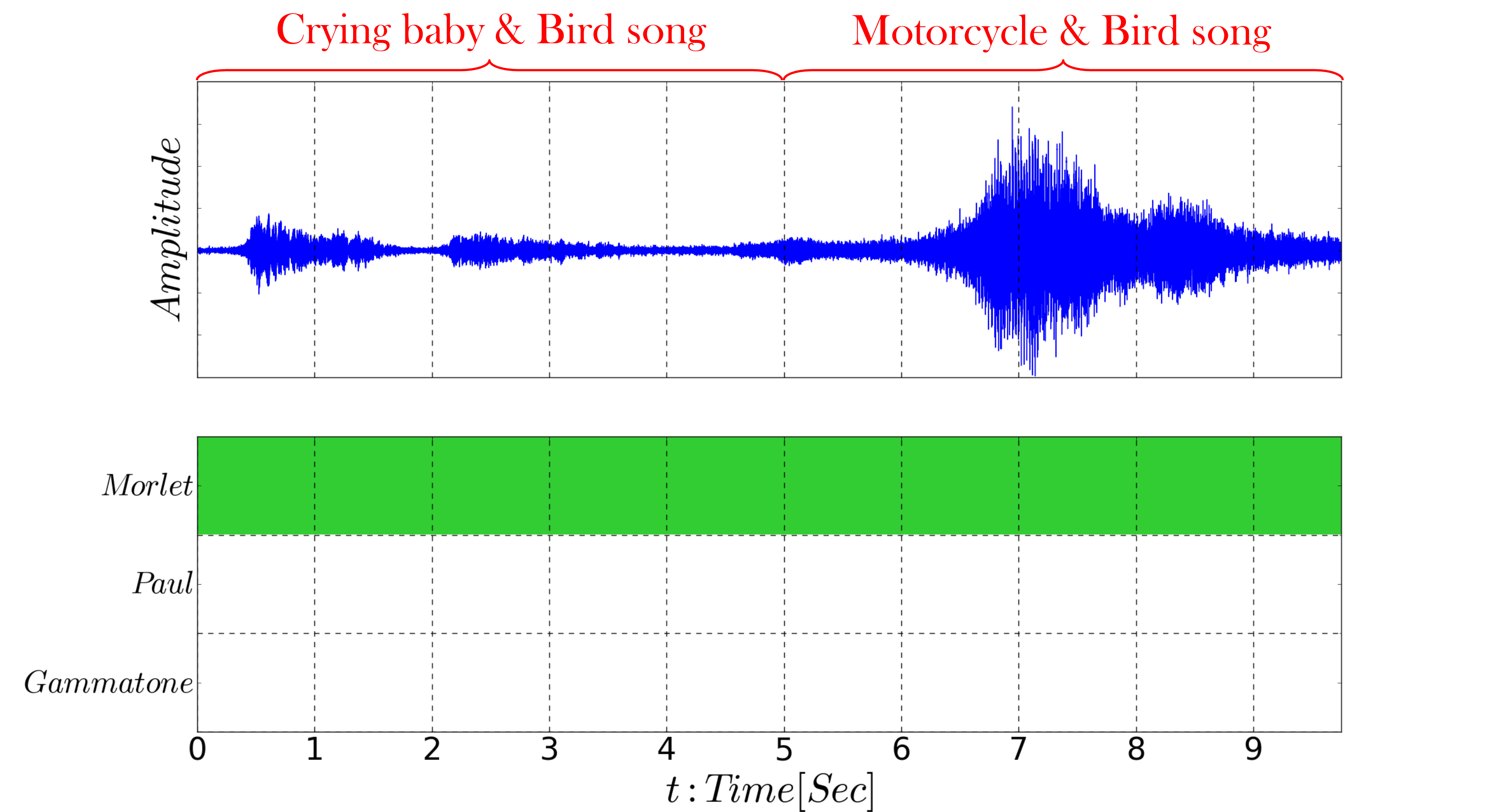}
    \caption{}
  \end{subfigure}
  \begin{subfigure}[b]{.5\textwidth}
    \centering
    \includegraphics[width=\textwidth]{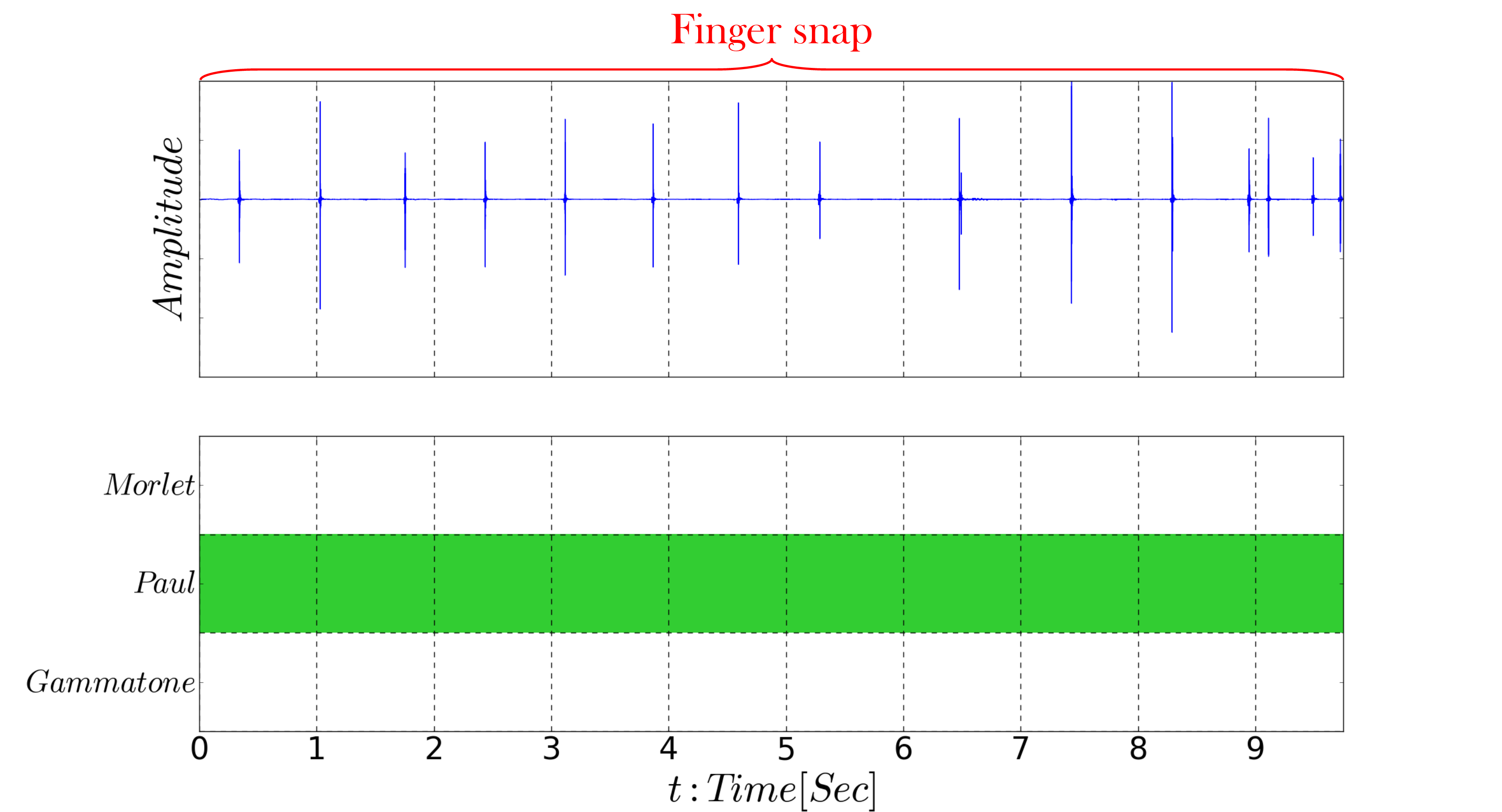}
    \caption{}
  \end{subfigure}
  \caption{\textbf{SMF-DSN Latent Wavelet Family Representation}: (a) Cocktail: Crying Baby \& Bird song \& Motorcycle - 15657.wav (c) Finger Snap Latent Representation - 349.wav.}
\label{fig:hidden_1}
\end{figure}

In Fig. \ref{fig:hidden_1}, we provide two samples from the dataset corresponding to very different acoustic scene. One represents transients on the right, while the left one provides a mixture of natural sounds. Risk-based analysis of the filter-banks fitness provides consistent information with the specificities of the selected wavelets. In fact, the Paul family is known to be particularly adapted for transient characterization via its high time localization \citep{torrence1998practical}. On the other hand, the Morlet wavelet is optimal with respect to the Heisenberg principle and thus suitable for natural sounds such as bird songs, speech, music.

\subsection{Validation: Bird Activity Detection benchmark} 
\label{validation}

We propose to validate the two contributions over a large-scale audio dataset. In all cases, the scattering coefficients are then fed into a random forest \citep{breiman2001random} with parameters\footnote{n\_estimator: $100$, criteria: 'gini', min\_samples\_split: from $50$ (small data size) to $150$ (all data),class\_weights:'balanced\_subsample'}  based on the sklearn library \cite{pedregosa2011scikit}

The data set consists of $4000$ field recording signals from  freefield1010\footnote{http://machine-listening.eecs.qmul.ac.uk/bird-audio-detection-challenge/} collected via the Freesound\footnote{https://arxiv.org/abs/1309.5275} project. This collection represents a wide range of audio scenes such as birdsong, city, nature, people, train, voice, water.
This paper focuses on the bird audio detection task that can be formally defined as a binary classification task, where each label corresponds to the presence or absence of birds.  Each signal is $10$sec. long, and has been sampled at $44.1$Khz. The evaluation of the results is performed via the Area Under Curve metric on $33\%$ of the data. The experiments are repeated 50 times.
This dataset's total audio length is thus slightly more than $11$ hours of audio recordings. To put in comparison, it is about $10 \times$ larger than CIFAR10 in terms of numbers of scalar values in the dataset. The results of SMF-DSN and MF-DSN are in Table \ref{table:compare}. For all the architectures, the octave and quality parameters of the layers are $J1=5,Q1=8,J2=4,Q2=1$. As the feature of interests is birds songs, only high-frequency content requires high resolution; the thresholding is applied per window of $2^{16}$ representing $\approx 1.5 sec$.

\begin{table*}[h!]
\centering
\caption{Classification Results - Bird Detection - Area Under Curve metric}
\begin{tabular}{llll}
\hline
\multicolumn{1}{l|}{}              & $\textbf{\textit{min}}$& \begin{tabular}[c]{@{}l@{}}$\textbf{\textit{mean}}$\end{tabular}& \begin{tabular}[c]{@{}l@{}}$\textbf{\textit{max}}$\end{tabular} \\ \hline
\multicolumn{1}{l|}{\textbf{Sparse Multi-Family Deep Scattering Network}}   \\ \hline  
\multicolumn{1}{l|}{$S_1$ $S_2$} & $\textbf{71.16}$  &$\textbf{73.52}$ & $\textbf{75.11}$   \\ \hline
\multicolumn{1}{l|}{\textbf{Multi-Family Deep Scattering Network}}   \\ \hline   
\multicolumn{1}{l|}{$S_1$ $S_2$} & $69.00$  &$71.17$ & $73.44$   \\
\end{tabular}
\label{table:compare}
\end{table*}

\section{Conclusion}

We presented an extension of the scattering network by composing and crossing various wavelet families in the DSN definition.  Such a topology exploits the specificity of the various wavelet families and removes the need for an expert with respect to the signal at hand.
We then proposed the analytical derivation of an optimal over-complete basis thresholding technique. Such input-dependant operation enables us to equip the DSN with a greater adaptive power while conserving its interpretability and stability. Besides, we showed that we could leverage the evaluation of the empirical risk as to understand which family characterized best the signal at hand. Finally, Combining these two techniques yields the Sparse Multi-Family Scattering Network, which conserves the DSN's benefits while extending its expressive power.


\bibliography{iclr2018_conference}
\bibliographystyle{iclr2018_conference}

\appendix

\section{Building a Multi-Family Deep Scattering Network}
\label{appendixA}
\subsection{Continous Wavelet Transform}\label{def}
\begin{center}
"By oscillating it resembles a wave, but by being localized it is a wavelet".\\ \raggedleft {Yves Meyer} 
\end{center}
Wavelets were first introduced for high resolution seismology \cite{cycle} \cite{grossmann1984decomposition} and then developed theoretically by Meyer et al. \cite{painless}.
Formally, wavelet is a function $\psi \in \mathbb{L}^2$ such that:
\begin{equation}
\int \psi(t) dt = 0,
\end{equation}
it is normalized such that $\left \| \psi \right \|_{\mathbb{L}^2} = 1$. 
There exist two categories of wavelets, the discrete wavelets and the continuous ones. The discrete wavelets transform are constructed based on a system of linear equation. These equations represent the atom's property. These wavelet when scaled in a dyadic fashion form an orthonormal atom dictionary.
Withal, the continuous wavelets have an explicit formulation and build an over-complete dictionary when successively scaled. In this work, we will focus on the continuous wavelets as they provide a more complete tool for analysis of signals. 
In order to perform a time-frequency transform of a signal, we first build a filter bank based on the mother wavelet. This wavelet is names the mother wavelet since it will be dilated and translated in order to create the filters that will constitute the filter bank. Notice that wavelets have a constant-Q property, thereby the ratio bandwidth to center frequency of the children wavelets are identical to the one of the mother. Then, the more the wavelet atom is high frequency the more it will be localized in time.
The usual dilation parameters follows a geometric progression and belongs to the following set:
\[ \Lambda = \left \{ 2^{j/Q}, j= 0,...,J \times Q -1 \right \}  \]. 
Where the integers $J$ and $Q$ denote respectively the number of octaves, and the number of wavelets per octave.
In order to develop a systematic and general principle to develop a filter bank for any wavelet family, we will consider the weighted version of the geometric progression mentioned above, that is:
\[ \Lambda = \left \{ \alpha 2^{j/Q}, j= 0,...,J \times Q -1 \right \}  \].
In fact, the implementation of wavelet filter bank can be delicate since the mother wavelet has to be define at a proper center frequency such that no artifact or redundant information will appear in the final representation. Thus, in the section \ref{filterbank} we propose a principled approach that allows the computation of the filter bank of any continuous wavelet. Beside, this re-normalized scaled is crucial to the comparison between different continuous wavelet.
Having selected a geometric progression ensemble, the dilated version of the mother wavelet in the time are computed as follows:   
    \[ \psi_{\lambda}(t)=\frac{1}{\lambda} \psi ( \frac{t}{\lambda} ),   \; \forall \lambda \in \Lambda       \],
and can be calculated in the Fourier domain as follows:
\[  \hat{\psi}_{\lambda}(\omega) = \hat{\psi}(\lambda \omega), \; \forall \lambda \in \Lambda            \]    .
    
Notice that in practice the wavelets are computed in the Fourier domain as the wavelet transform will be based on a convolution operation which can be achieved with more efficiency. 
By construction the children wavelets have the same properties than the mother one. As a result, in the Fourier domain:
\[ \hat{\psi}_{\lambda} = 0, \; \forall \lambda \in \Lambda \].
Thus, to create a filter bank that cover all the frequency support, one needs a function that captures the low frequencies contents. The function is called the scaling function and satisfies the following criteria:
\[ \int \phi(t) dt = 1 \].


Finally, we denote by $Wx$, where $W \in \mathbb{C}^{N*(JQ) \times N}$ is a block matrix such that each block corresponds to the filters at all scales for a given time. Also, we denote by $S(Wx)(\lambda,t)$ the reshape operator such that,
\begin{equation}
S(Wx)(\lambda,t) = (\frac{1}{\sqrt\lambda}\psi^{\star}_{\lambda}\star x)(t), \forall \lambda \in \Lambda,
\end{equation}
where $\psi^{\star}$ is the complex conjugate of $\psi_{\lambda}$.
\subsection{Wavelet Families}
\label{family}
Among the continuous wavelets, different selection of mother wavelet is possible. Each one posses different properties, such as bandwidth, center frequency. This section is dedicated to the development of the families that are important for the analysis of diverse signals.
\subsubsection{The Morlet wavelet}
\begin{figure}[H]
    \centering
    \includegraphics[width=.9\linewidth]{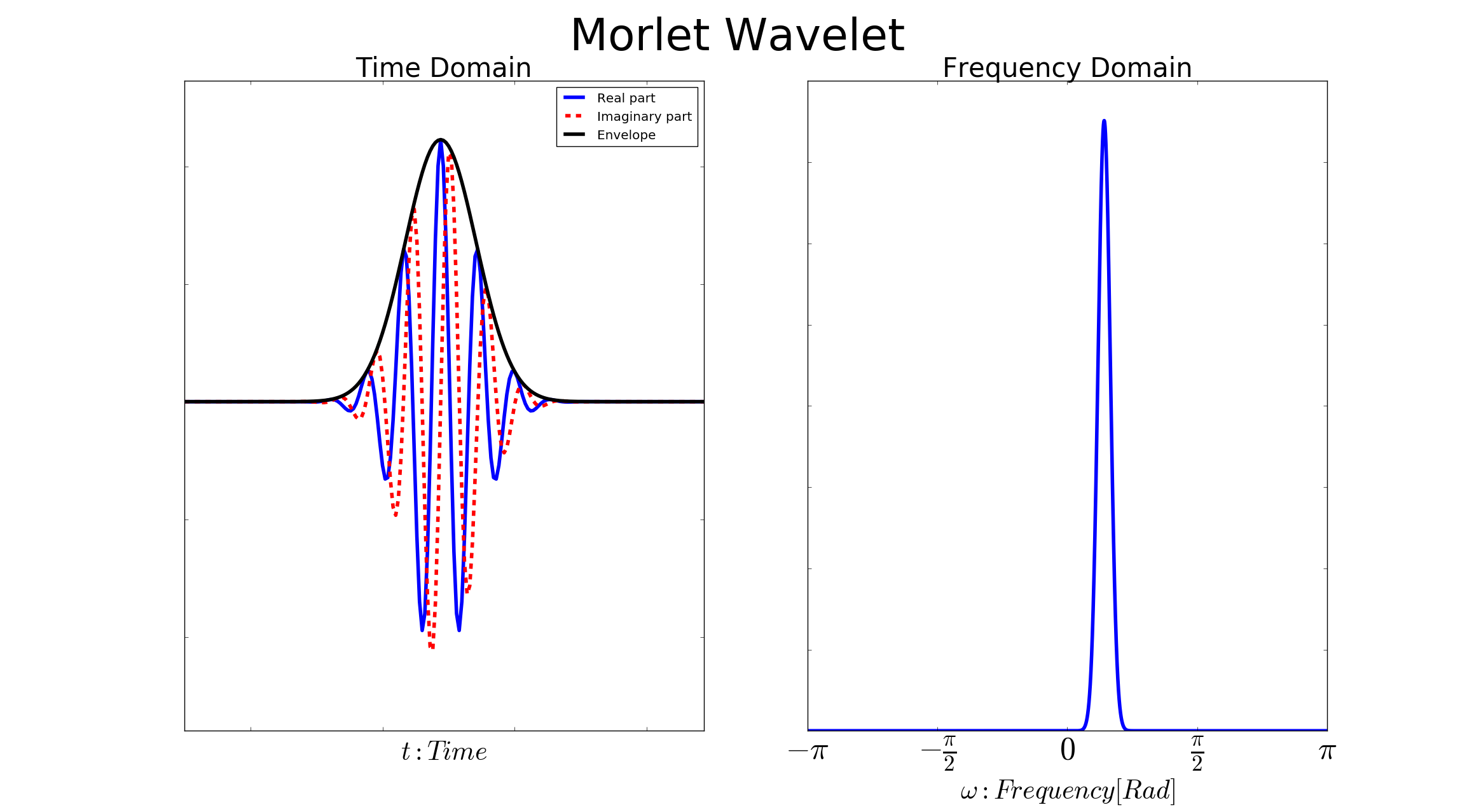}
    \caption{On the left a Morlet wavelet in the time domain where the dashed line is the imaginary part, the solid line is the real part, and the black envelope is the complex modulus, on the right a Morlet wavelet in the frequency domain.}
    \label{fig:Morlet}
\end{figure}

The Morlet wavelet (Fig. \ref{fig:Morlet}) is built by modulating a complex exponential and a Gaussian window defined in the time domain by,
\begin{equation}
\label{Morlet}
\psi^{\textit{M}}(t) = \pi^{-\frac{1}{4}} e^{i\omega_0 t}e^{-\frac{t^2}{2}},
\end{equation}
where $\omega_0$ defines the frequency plane.
In the frequency domain, we denote it by $\hat{\psi}^{\textit{M}}(t)$,
\begin{equation}
\hat{\psi}^{\textit{M}}(\omega) = \pi^{-\frac{1}{4}}e^{-\frac{(\omega-\omega_0)^2}{2}}, \forall \omega \in \mathbb{R}^{\star}_{+},
\end{equation}
thus, it is clear that $\omega_0$ defines the center frequency of the mother wavelet.

With associated frequency center and standard deviation denoted respectively by $\omega_{c}^{\lambda_i}$ and $\Delta^{\lambda_i}\omega$, $\forall j \in \{0,...,JQ-1 \}$ are:
\begin{align*}
    \omega_{c}^{\lambda_i}&= \frac{\omega_0}{\lambda_i} \nonumber, \\ 
    \Delta^{\lambda_i}\omega&= \frac{1}{2 \lambda_i^2} \nonumber.
\end{align*}
Notice that for the admissibility criteria $\omega_0 = 6$, however one can impose that zeros-mean condition facilely in the Fourier domain. Usually, this parameter is assign to the control of the center frequency of the mother wavelet, however in our case, we will see in the section \ref{filterbank} a simple way to select a mother wavelet close enough to the Nyquist frequency such that all its contracted versions are properly defined. Then, we are able to vary the parameter $\omega_0$ in order to have different support of Morlet wavelet. 
The Morlet wavelet, is optimal from the uncertainty principle point of view \cite{Wavtour}. The uncertainty principle, when given a time-frequency atoms, is the area of the rectangle of its joint time-frequency resolution. In the case of wavelet, given the fact that their ratio bandwidth to center frequency is equal implies that this area is equal for the mother wavelets and its scaled versions. As a result, because of its time-frequency versatility this wavelet is wildly used for biological signals such as bio-acoustic \cite{balestriero2015scattering}, seismic traces \cite{chopra2015choice}, EEG \cite{dwavelet} data.

\subsubsection{The Gammatone wavelet}
\begin{figure}[H]
\centering
\begin{subfigure}[b]{.8\textwidth}
    \includegraphics[width=.9\textwidth]{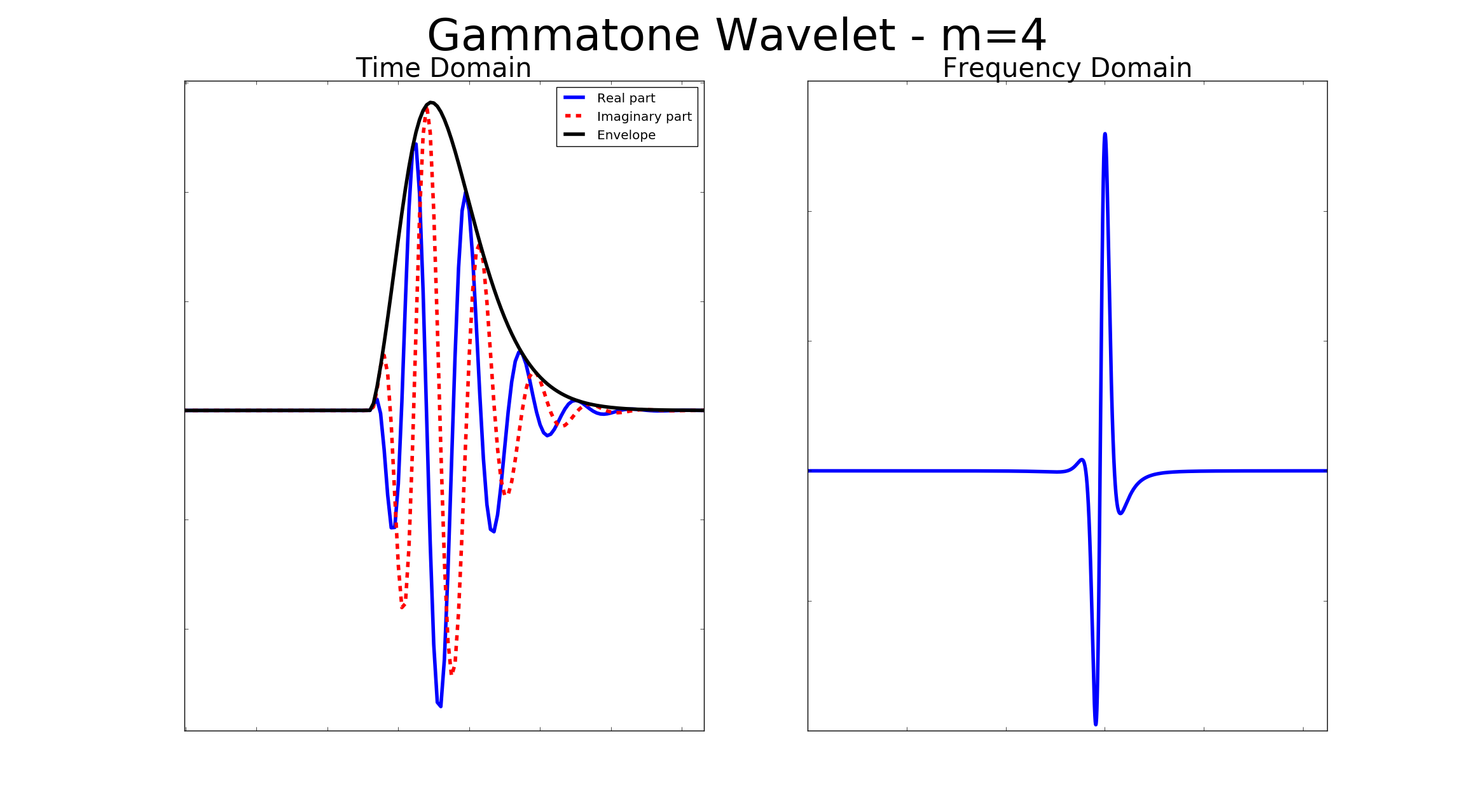}
\end{subfigure}
~
\begin{subfigure}[b]{.8\textwidth}
    \includegraphics[width=.9\textwidth]{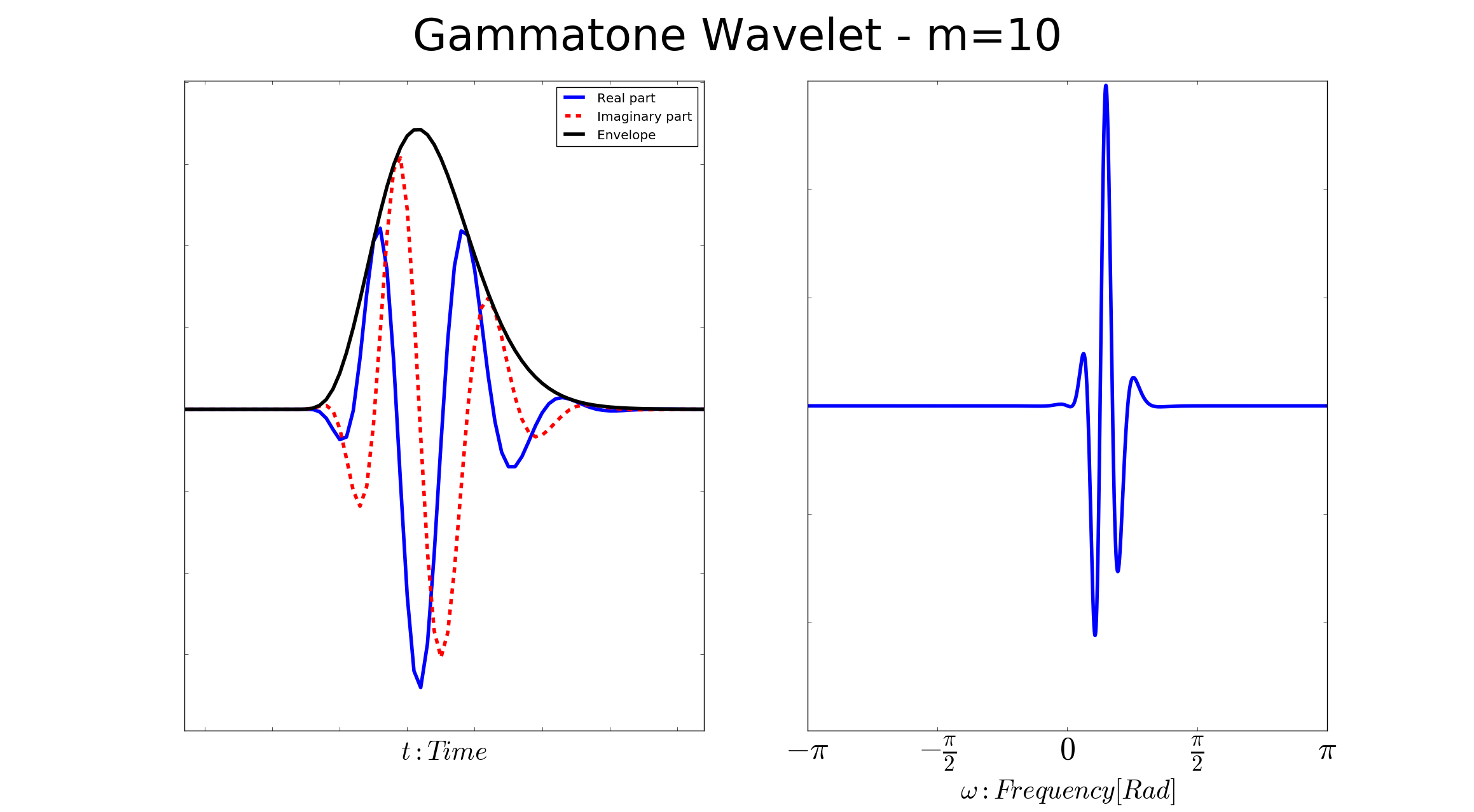}
\end{subfigure}
\caption{On the upper (bottom) left a $m=4$ ($m=10$) Gammatone wavelet in the time domain where the dashed line is the imaginary part and the solid line is the real part, on the upper (bottom) right a $m=4$  ($m=10$) wavelet in the frequency domain.}\label{fig:Gammatone}
\end{figure}

The Gammatone wavelet is a complex-valued wavelet that has been developed by \cite{venkitaraman2014auditory} via a transformation of the real-valued Gammatone auditory filter which provides a good approximation of the basilar membrane filter \cite{flanagan1960models}. Because of its origin and properties, this wavelet has been successfully applied for classification of acoustic scene \cite{lostanlenbinaural}.
The Gammatone wavelet (Fig. \ref{fig:Gammatone}) is defined in the time domain by,
\begin{equation}
\label{Gammatone}
\psi^{\textit{G}}(t) = \left ( 2 \pi(i-\sigma)t^{m-1}+(m-1)t^{m-2}\right )e^{-2pi\sigma t}e^{2 pi i t},
\end{equation}
and in the frequency domain by,
\begin{equation}
\hat{\psi}^{\textit{G}}(\omega) =  \frac{i \omega (m-1)!}{\left (\sigma +i(\omega-\sigma) \right )^m}.
\end{equation}
A precise work on this wavelet achieved by V. Lostalnen in \cite{lostanlen2017operateurs} allows us to have an explicit formulation of the parameter $\sigma$ such that the wavelet can be scaled while respecting the admissibility criteria:
\begin{align*}
\sigma^2 = \frac{ r^{\frac{2}{m}} (1 - r^{\frac{2}{m}}) m^2 \xi^2}{2} \left ( \sqrt{1+ \frac{B^2}{(1 - r^{\frac{2}{m}})^2 m^2 \xi^2}} -1 \right ),
\end{align*}
where $\xi$ is the center frequency and $B$ is the bandwidth parameter. Notice that $B=(1-2^{-\frac{1}{Q}}) \xi$ with $\xi= \frac{2 \pi}{1+ 2^{\frac{1}{Q}}}$ induce a quasi orthogonal filter bank.
The associated frequency center and standard deviation denoted respectively by $\omega_{c}^{\lambda_i}$ and $\Delta^{\lambda_i}\omega$, $\forall j \in \{0,...,JQ-1 \}$ are thus:
\begin{align*}
    \omega_{c}^{\lambda_i}&= \xi \nonumber, \\ 
    \Delta^{\lambda_i}\omega&= B \nonumber.
\end{align*}
For this wavelet, thanks to the derivation in \cite{lostanlen2017operateurs}, we can manually select for each order $m$ the center frequency and bandwidth of the mother wavelet, which ease the filter bank design.
An important property that is directly related to the auditory response system is the asymmetric envelop, thereby the Gammatone wavelet is not invariant to time reversal to the contrary of the Morlet wavelet that behaves as a Gaussian function. Thus, for task such as sound classifications, this wavelet provides an efficient filter that will be prone to perceive the sound attack's. Beside this suitable property for specific analysis, this wavelet is near optimal with respect to the uncertainty principle. Notice that, when $m \rightarrow \infty$ it yields the Gabor wavelet \cite{Cohen:1995}. Another interesting property of this wavelet is the causality, by taking into account only the previous and present information, there is no bias implied by some future information and thus it is suitable for real time signal analysis.

\subsubsection{The Paul wavelet}
\begin{figure}[H]
\centering
\begin{subfigure}[b]{.8\textwidth}
    \includegraphics[width=.9\textwidth]{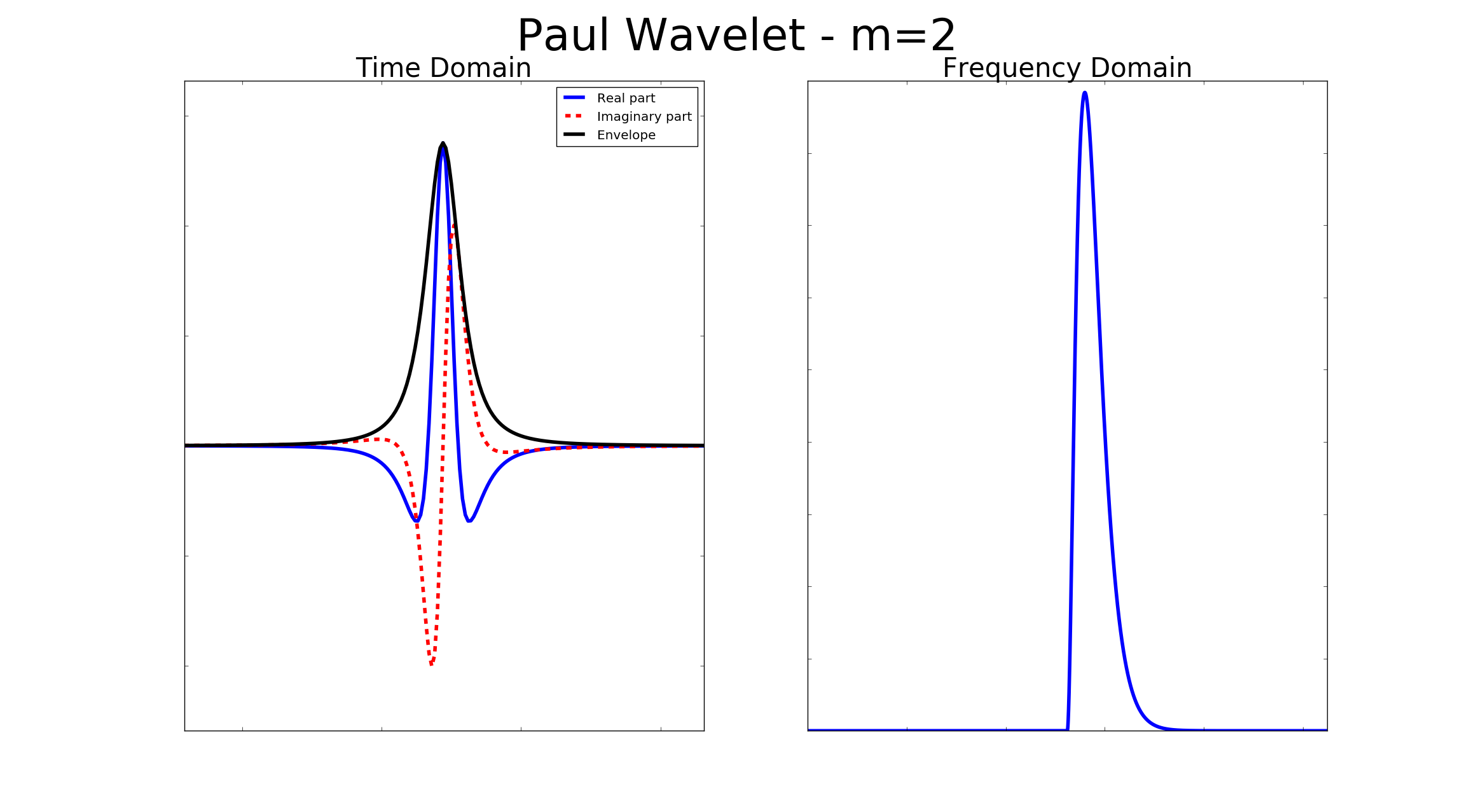}
\end{subfigure}
~
\begin{subfigure}[b]{.8\textwidth}
    \includegraphics[width=.9\textwidth]{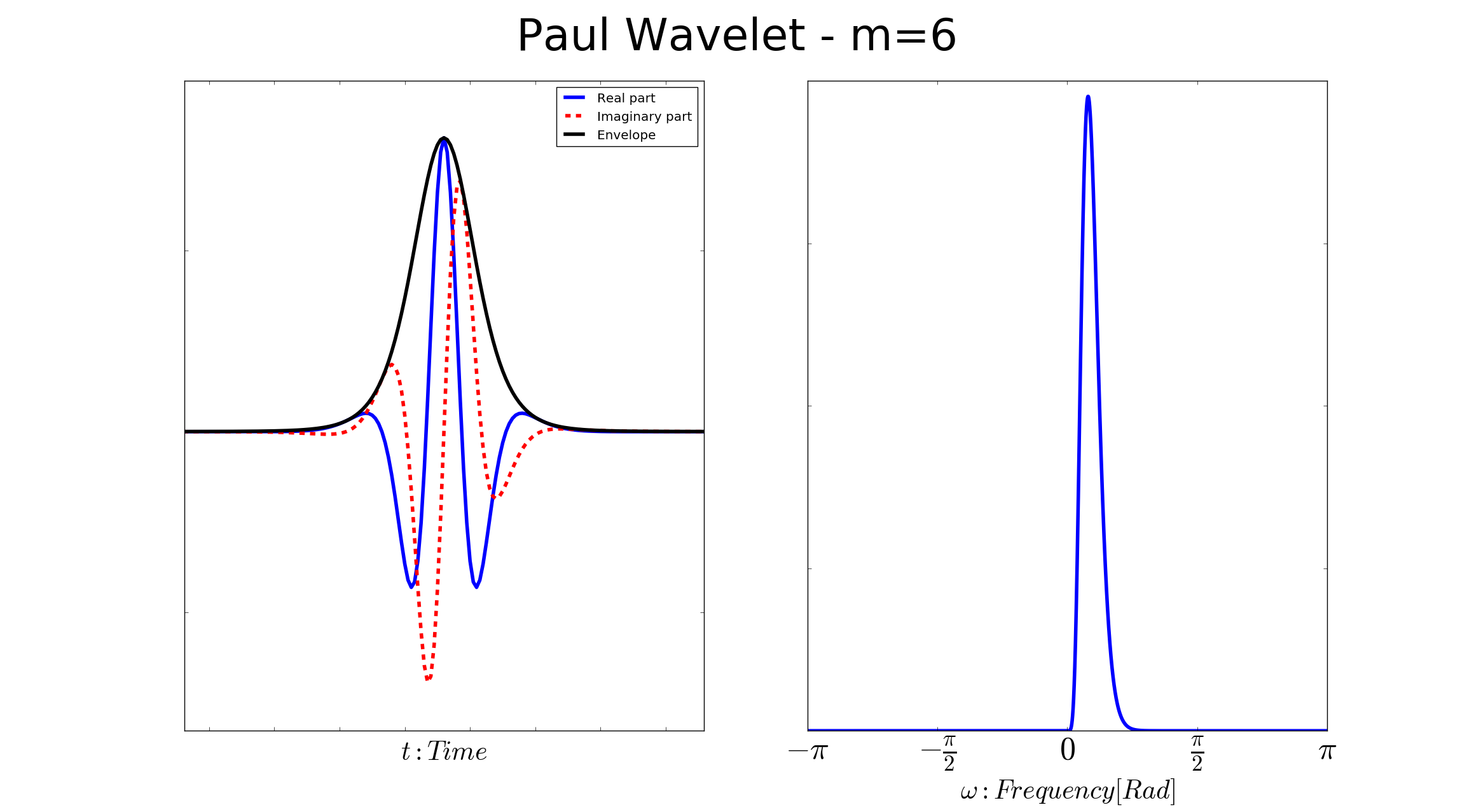}
\end{subfigure}
\caption{On the upper (bottom) left a $m=2$ ($m=6$) Paul wavelet in the time domain where the dashed line is the imaginary part and the solid line is the real part, on the upper (bottom) right a $m=2$  ($m=6$) wavelet in the frequency domain.}\label{fig:Paul}
\end{figure}

The Paul wavelet is a complex-valued wavelet which is highly localized in time, thereby has a poor frequency resolution. Because of its precision in the time domain, this wavelet is an ideal candidate to perform transient detection. 
The Paul wavelet of order $m$ ( Fig. \ref{fig:Paul}) is defined in the time domain by,
\begin{equation}
\psi^{\textit{P}}(t) = \frac{2^m i^m m!}{\sqrt{2m! \pi}} (1-i t)^{-(m+1)}
\end{equation}
and in the frequency domain by,
\begin{equation}
\hat{\psi}^{\textit{P}}(t) = \frac{2^m}{\sqrt{m(2m-1)!}} (\omega)^{m} e^{-\omega}, \forall \omega \in \mathbb{R}^{\star}_{+},
\end{equation}
With associated frequency center and standard deviation denoted respectively by $\omega_{c}^{\lambda_i}$ and $\Delta^{\lambda_i}\omega$ $, \forall j \in \{0,...,JQ-1 \}$ are:
\begin{align*}
    \omega_{c}^{\lambda_j}&= \frac{2m+1}{2 \lambda_j} \nonumber, \\ 
    \Delta^{\lambda_j}\omega&= \frac{\sqrt{2m +1}}{2 \lambda_j} \nonumber.
\end{align*}
In \cite{torrence1998practical} they provide a clear and explicit formulation of some wavelet families applied the Paul wavelet in order to capture irregularly periodical variation in winds and sea surface temperatures over the tropical eastern Pacific Ocean . In addition, it directly represents the phase gradient from a single fringe pattern, yet providing a powerful tool in order to perform optical phase evaluation  \cite{afifi2002paul}.

\subsection{Filter Bank Design}
\label{filterbank}
\begin{figure}[H]
    \centering
    \includegraphics[width=.9\linewidth]{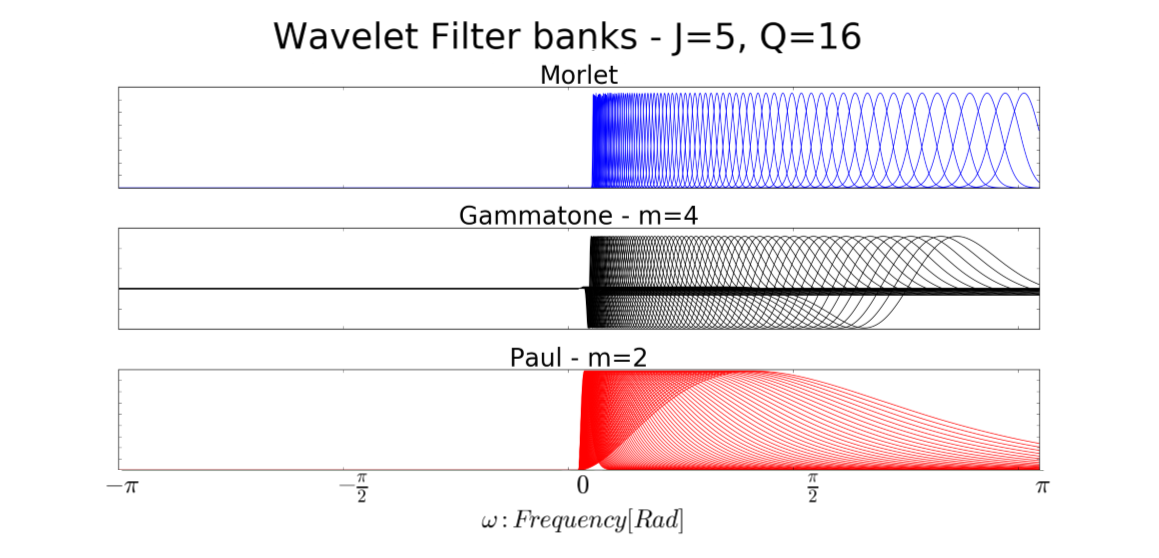}
    \caption{\textbf{From top to bottom}: Morlet wavelet filter bank, Gammatone wavelet filter bank with $m=4$, Paul wavelet filter bank $m=2$}
    \label{fig:filterbank}
\end{figure}

In the previous section, we defined and develop the properties of several families of wavelets. Thereby, we can now consider the creation of the filter bank by means of these wavelets. Notice that we propose a simple manner to obtain the filter bank in the Fourier domain. Two reasons are at the origin of this choice: first, the wavelet transform is often computed in the Fourier domain because of its efficiency, secondly the wavelets are derived according to geometric progression scales, these scales can directly be represented in the frequency domain, thereby it provided us a way of knowing the position of the wavelet. However, in the time domain they are not directly quantifiable. Our systematic framework is based on the intuitive consideration of the problem: we have to select a wavelet, named mother wavelet, that when contracted will create the filter bank derived from the selected scales. Assuming that the signals we will use are real valued,  then the information represented in $\left [-\pi,0 \right ]$  and $\left [0,\pi \right ]$ are the same if extracted with a symmetric atom. Now, two kind of wavelets are considered, if the wavelet is complex-valued then its support is in $\left [0,\pi \right ]$, thus the choice of the mother wavelet should be around $\pi$ and the contracted all along the frequency axis until the total number of octave are covered. In the case of real-valued wavelet, if the wavelet is not symmetric then it will capture other phase information in the frequency band:$\left [-\pi,0 \right ]$. Still, the mother wavelet can be selected to be close to $\pi$ for its positive part, and $-\pi$ for its negative one.
After defining the routine in order to select the mother wavelet, we propose a simple way to set the position of the mother wavelet. For each family, the center frequency and standard deviation are derived by finding $\alpha$ such that:
\begin{equation}
     \omega_c^{\lambda_0} + \Delta^{\lambda_0} \omega = \pi ,
\end{equation}     
where $\lambda_0 = \alpha *2^{0/Q}$ denotes the first wavelet position. Given this equation, one create the mother wavelet such that it avoids capturing elements after the Nyquist frequency and avert the spectral mirror effect and artifacts. Given the value of $\alpha$ for a wavelet family, one can derive the wavelet filter bank according to the Algorithm \ref{algo:filterbank}. The wavelet filter banks generated by this algorithm for the different families aforementioned can be seen in Fig. \ref{fig:filterbank}. Notice that for sake of clarity, the scaling functions are not shown in Fig. \ref{fig:filterbank}. 
\begin{algorithm}[H]
  \SetAlgoLined
\KwIn{wavelet family: $f \in \mathcal{F}$, signal length: $N$, number of octaves: $J$, number of wavelets per octave: $Q$, scale weight: $\alpha$, wavelet parameter: $m$}
 \KwOut{$D:$ Filter bank of the wavelet family $f$ }
Initialize the wavelet frequency domain: $\omega \in \left [-\pi,\pi  \right ]$\\
 \While{$ j< JQ$}{
$\lambda_j = \alpha^{f}2^{\frac{j}{Q}}$ - Set up the scale for the $jth$ children wavelet - \\
$D_j= \psi^f(\lambda_j*\omega)$  - Compute the children wavelet at the given scale $\lambda_j$ -\\ 
 }
 \caption{Compute Filter bank for any continuous wavelet family $f \in \mathcal{F}$}\label{algo:filterbank}
\end{algorithm}

Finally, in order to guarantee the admissibility criterion one has to verify that all the wavelets are zeros-mean and square norm one. The first one is easily imposed by setting the wavelet to be null around $\omega=0$ as it has been done to efficiently use the Morlet wavelet by Antoine et. al   \cite{farge1992wavelet,antoine1993image}. Then, because of Parseval equality and the energy conservation principle, the second one can be achieved by a re-normalization in the frequency domain of each atom. 

\section{Activation Histogram: Sparsity Evaluation Layer 2}
    \label{appendixB}
    \begin{figure}[H]
        \centering
        \begin{subfigure}[b]{0.475\textwidth}
            \centering
            \includegraphics[width=\textwidth]{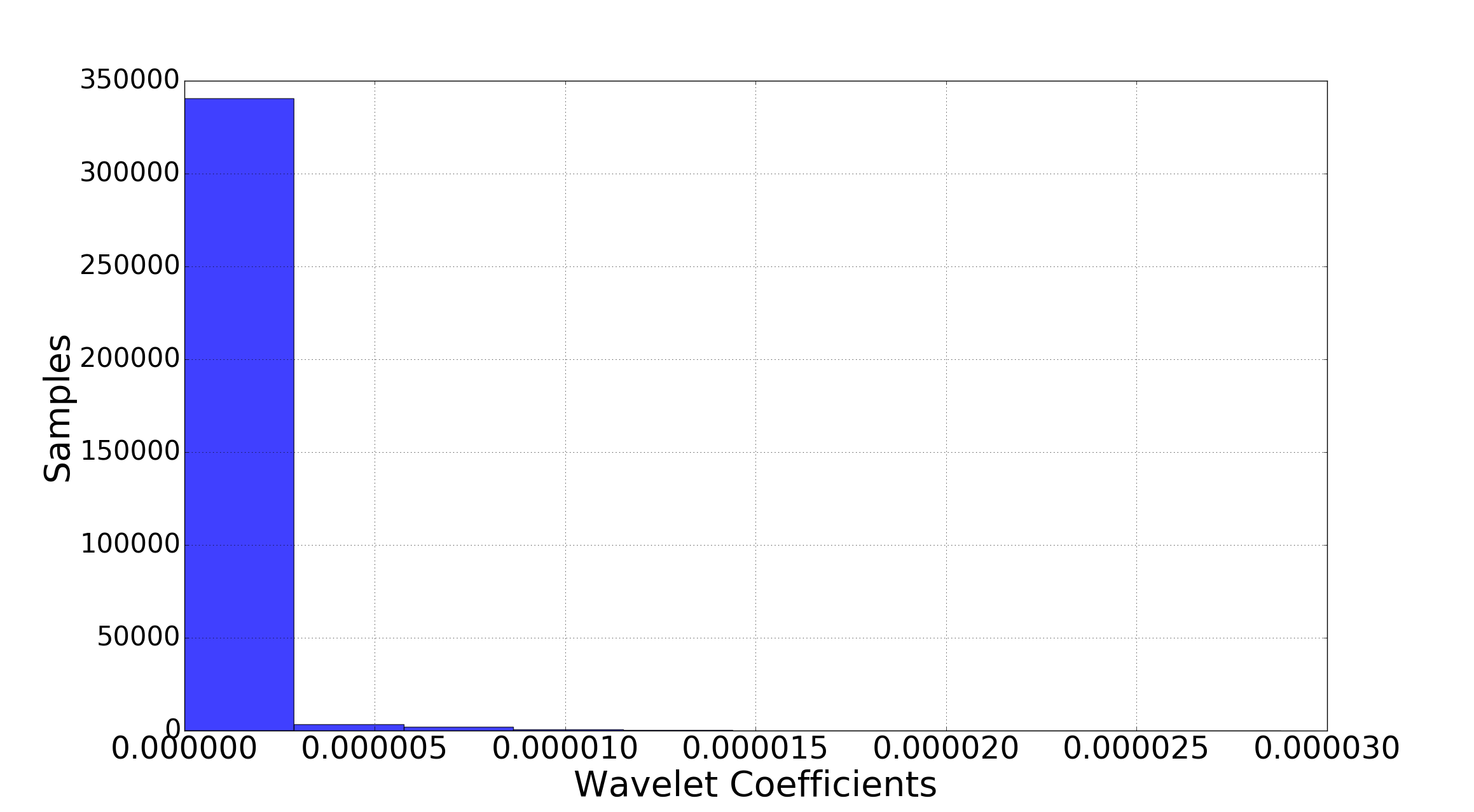}
            \caption{SCSN 2nd Layer\\Gammatone $\rightarrow$ Gammatone} 
        \end{subfigure}
        \quad
        \begin{subfigure}[b]{0.475\textwidth}  
            \centering 
            \includegraphics[width=\textwidth]{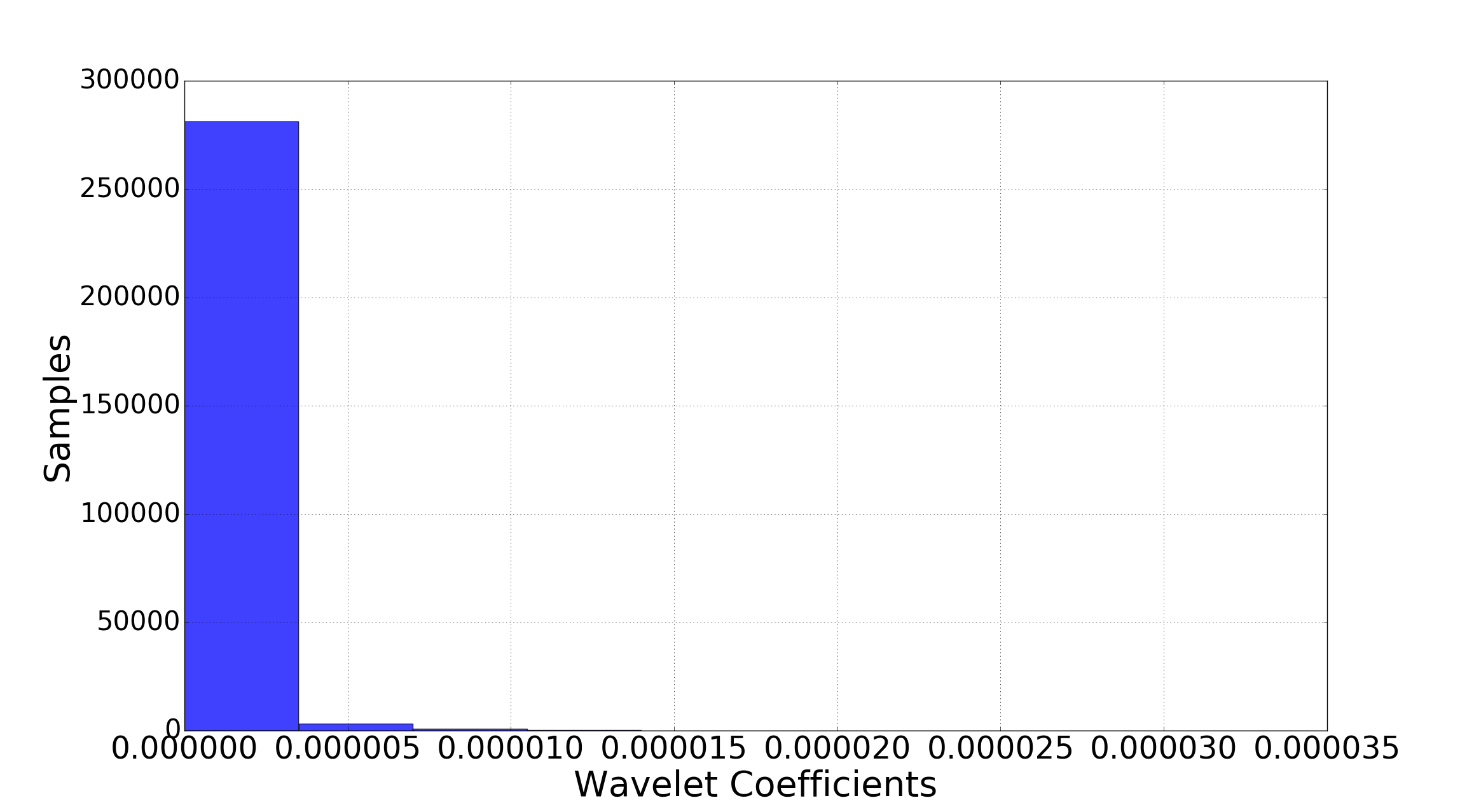}
            \caption{CSN  2nd Layer \\ Gammatone $\rightarrow$ Gammatone}     
        \end{subfigure}
        \vskip\baselineskip
        \begin{subfigure}[b]{0.475\textwidth}   
            \centering 
            \includegraphics[width=\textwidth]{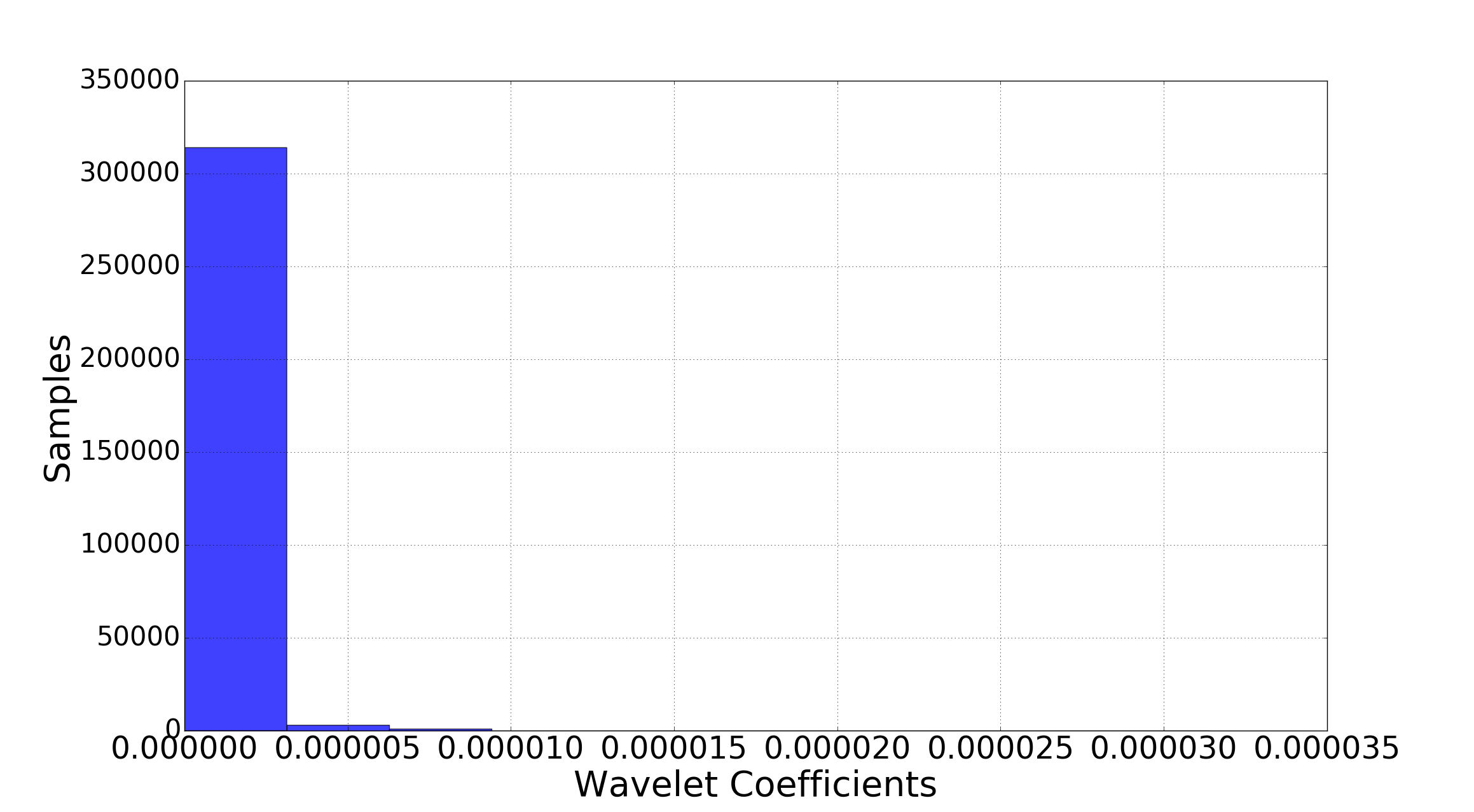}
            \caption{SCSN 2nd layer\\ Gammatone $\rightarrow$ Morlet }    
        \end{subfigure}
        \quad
        \begin{subfigure}[b]{0.475\textwidth}   
            \centering 
            \includegraphics[width=\textwidth]{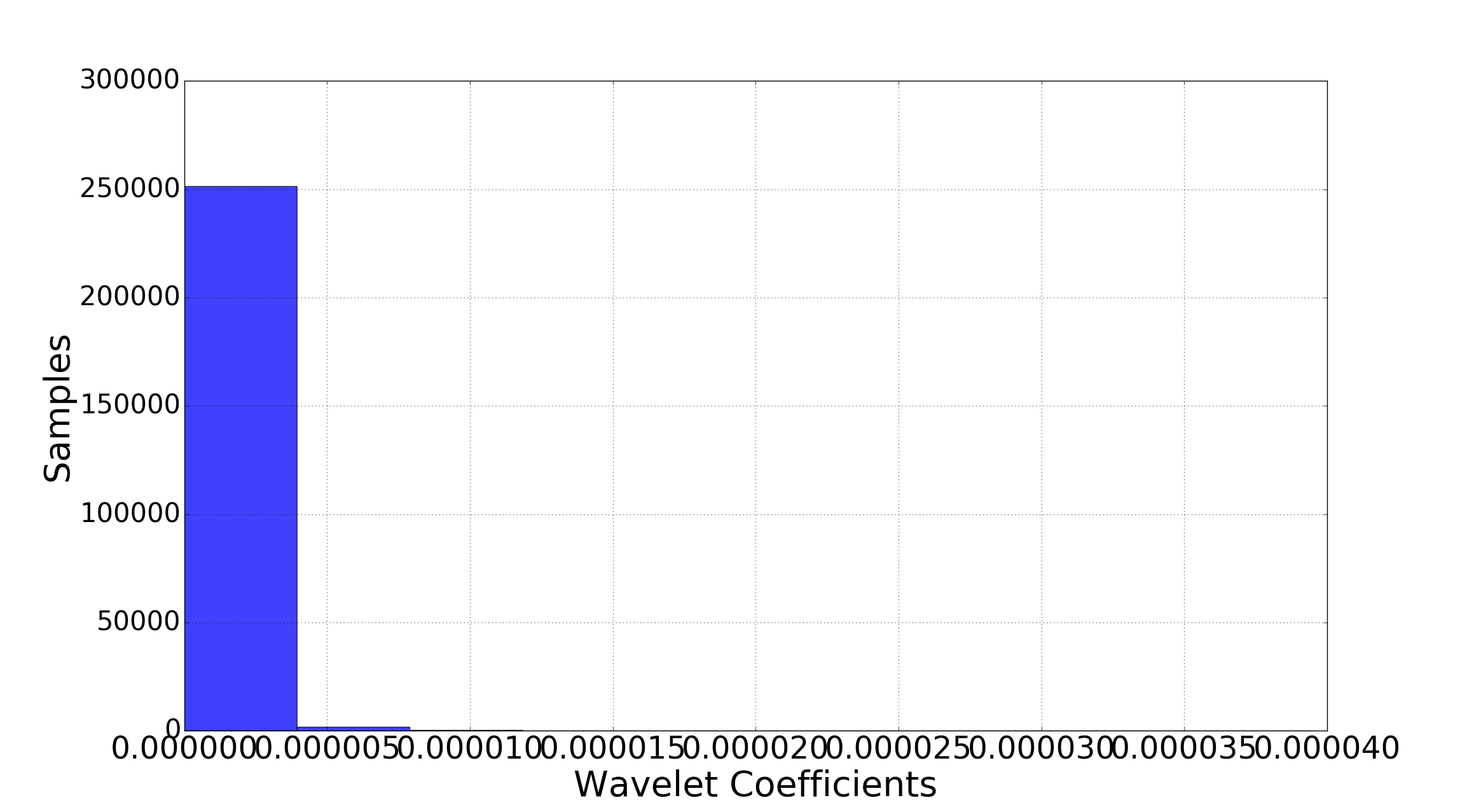}
            \caption{CSN  2nd Layer\\ Gammatone $\rightarrow$ Morlet}    
        \end{subfigure}
                \vskip\baselineskip
        \begin{subfigure}[b]{0.475\textwidth}   
            \centering 
            \includegraphics[width=\textwidth]{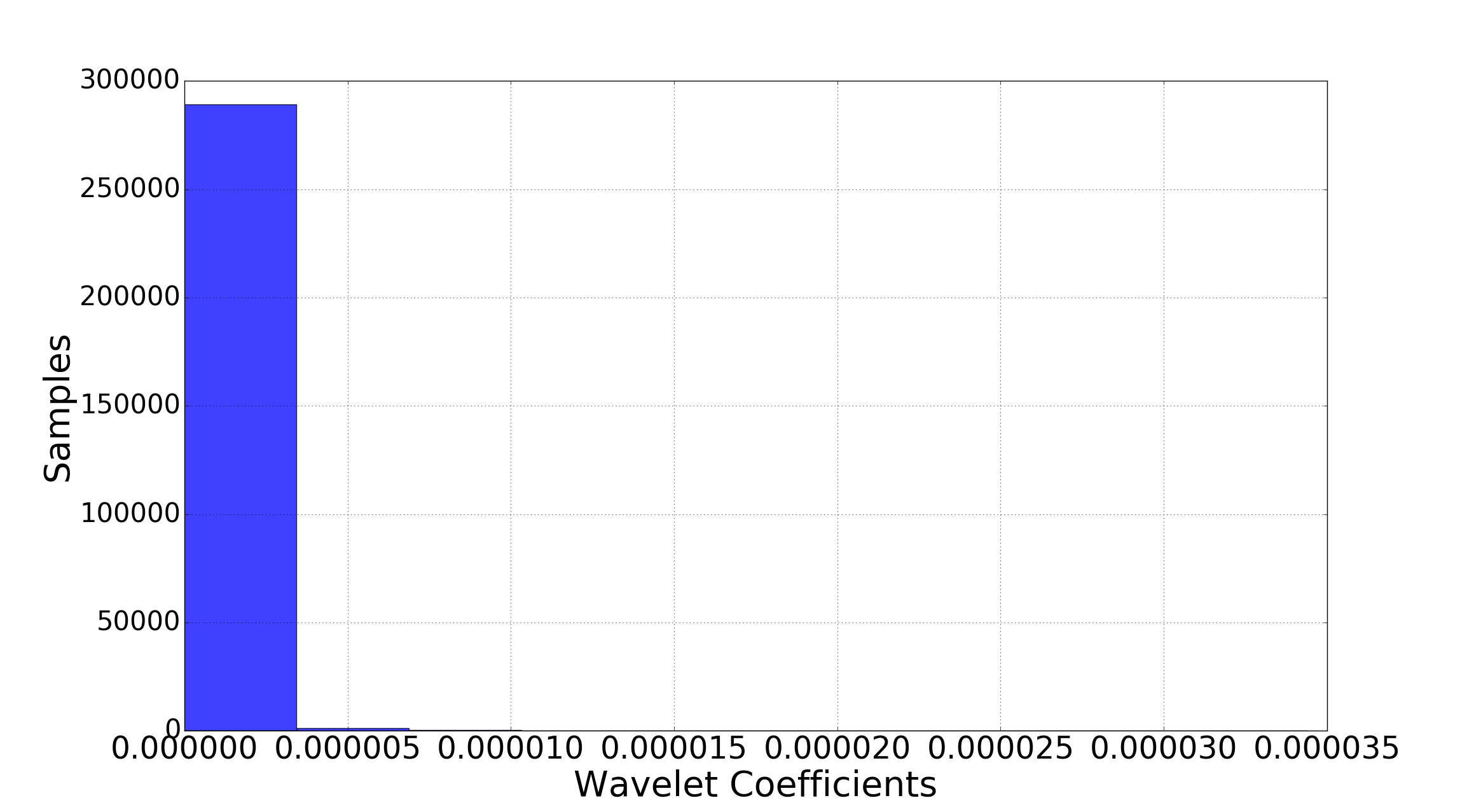}
            \caption{SCSN 2nd Layer\\ Gammatone $\rightarrow$ Paul}     
        \end{subfigure}
        \quad
        \begin{subfigure}[b]{0.475\textwidth}   
            \centering 
            \includegraphics[width=\textwidth]{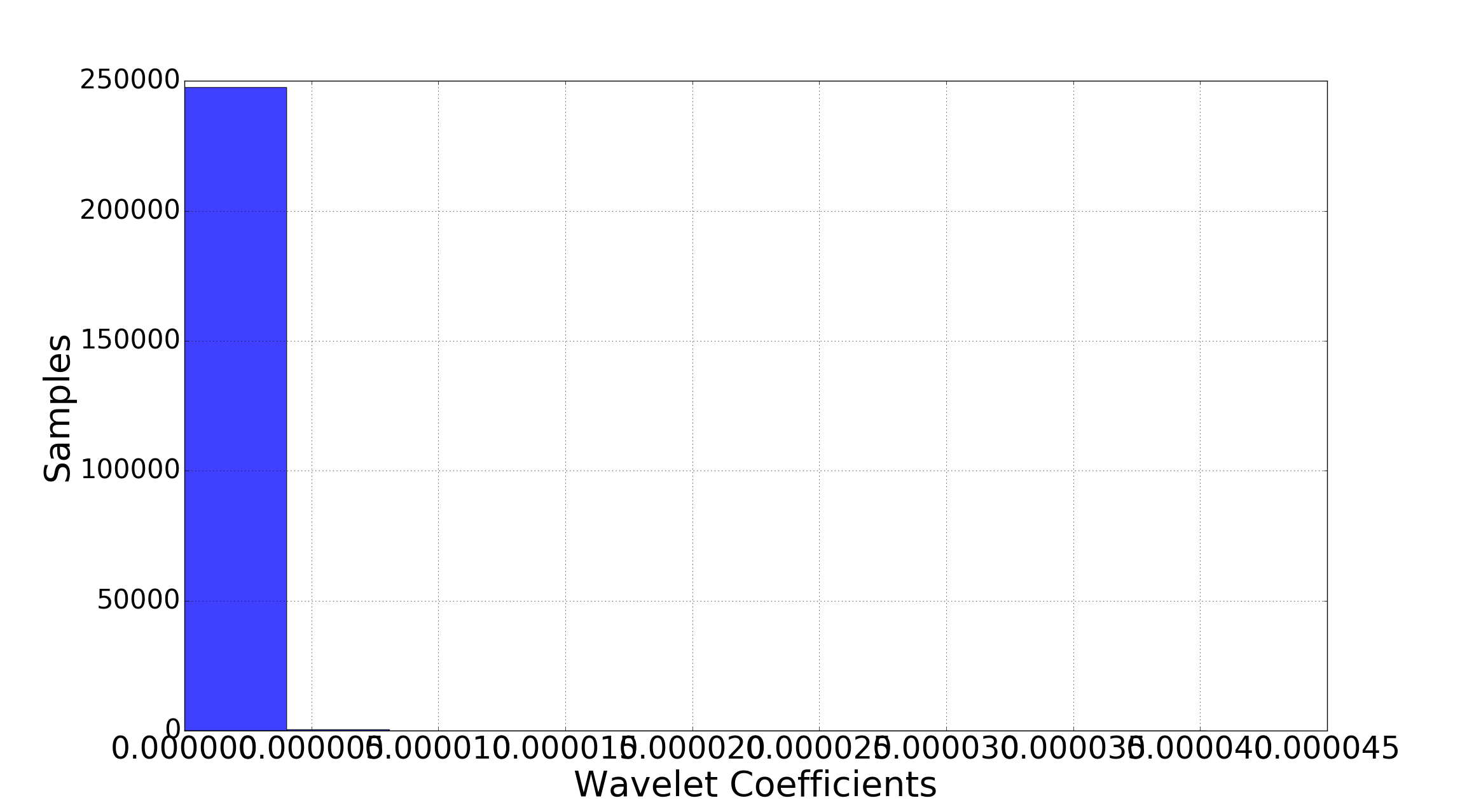}
            \caption{CSN  2nd Layer\\ Gammatone $\rightarrow$ Paul}     
        \end{subfigure}
        \caption{\textbf{Histogram of activation of the second layer activations}}
        \label{fig:histo-activation-2nd}
    \end{figure}

\section{Denoising in an orthogonal basis framework}
\label{appendixC}

Assuming that the observed signal $y$, is corrupted with white noise, 
 \begin{equation}
 y = x+ \epsilon,
 \end{equation}
 where $\epsilon$ is a vector of i.i.d centered normal distributions $\mathcal{N}(0,\sigma^2)$. 
 Now we define the estimate of $x$ by $\hat{x}_{W,D}$ such that:
 \begin{equation}
     \hat{x}_{W,D}(y) = W^{T}D^{S}Wy
 \end{equation}
 where $W$ denotes the orthogonal basis and $D^{S}$ is a diagonal binary operator such that,
 \begin{equation}
 D^{S}_{i,i}= \delta_i = \left\{\begin{matrix}
 1  \: \: if \: \: i \in S  \\ 
 0  \: \: if \: \: i \in U 
\end{matrix}\right. 
\end{equation},
 where $U$ and $S$ denote respectively the set of selected and unselected wavelet coefficients. We also define $D^{U}$ such that $I = D^{U} + D^{S}$.
 This estimate corresponds to a thresholding operation in the new basis and the inverse transform of this truncated representation. 

 We define the denoising problem as the solution of the following mean-square error:
 \begin{align}
 \mathcal{R}_{o}^{\star}(x,W)&=  \min_{\delta} \mathbb{E} \left \| x-\hat{x}_{W,D}(y) \right \|^2 \\
 & =   \min_{\delta} \mathbb{E} \left \| W^{T} (Wx-D^{S}Wy \right \|^2 \\
 & =  \min_{\delta} \mathbb{E} \left \|D^{U} Wx - D^{S}W(x+\epsilon) \right \|^2 \\
 & =  \min_{\delta} \left \| D^{U} Wx  \right \|^2 + \sigma^{2} tr (D^{S} WW^{T} D^{S}) \\
 & =  \min_{\delta} \sum_{i}^{n} [Wx]_i^{2} 1_{\left \{ \delta_i = 0 \right \}} + \sigma^{2} 1_{\left \{  \delta_i =1\right \}} \\
 & =  \sum_i^{n} \min ([Wx]_{i}^{2},\sigma^2).
 \end{align}
 Therefore, the optimal $D^{S^{\star}}$ and $D^{U^{\star}}$ given by the following $\delta$ values:
\begin{equation}
   \delta_i =  1_{\left | [Wx]^2_i \right |>\sigma^2}.
\end{equation}
\subsection{Upper-bound Non-orthogonal Risk \& Empirical Risk}
\label{upper-bound}

\begin{equation} 
     \hat{x}_{W,D}(y) = W^{\dagger}D^{S}Wy
 \end{equation}
\begin{align}
 \mathcal{R}^{\star}(x,W)&=  \min_{\delta} \mathbb{E} \left \| x-\hat{x}_{W,D}(y) \right \|^2 \\
 & =   \min_{\delta} \mathbb{E} \left \| W^{\dagger} (Wx-D^{S}Wy \right \|^2 \\
 & =  \min_{\delta} \mathbb{E} \left \|W^{\dagger} (D^{U} Wx - D^{S}W\epsilon) \right \|^2 \\
 & =  \min_{\delta} \left \| W^{\star}D^{U} Wx  \right \|^2 + \sigma^{2} tr (W^{T}D^{S} W^{\dagger}W^{\dagger^{T}} D^{S}W),
 \end{align}
Developing the previous expression and denoting by $\mu =  Wx$ the wavelet coefficient vector, we have:
\begin{align}
 \mathcal{R}^{\star}(x,W) = & \min_{\delta} \sum_{t=1}^{n} \sum_{i,j=1}^{N(JQ+1)} \mu_{i}\mu{j} \psi^{\dagger}_{t}[i] \psi^{\dagger}_{t}[j] 1_{\left \{ \delta_i = 0, \delta_j = 0 \right \} } \nonumber \\
 & + \sigma^{2} \sum_{i,j=1}^{N(JQ+1)}  ( \sum_{t=1}^{n} \psi^{\dagger}_{t}[i] \psi^{\dagger}_{t}[j] ) \psi_i^{T} \psi_j 1_{\left \{ \delta_i =1,\delta_j =1 \right \}}.
 \end{align}
we first use the triangular inequality,

\begin{align}
    \mathcal{R}^{\star}(x,W)  & \leq  \min_{\delta} \sum_{i,j=1}^{N(JQ+1)}  \left |  \sum_{t=1}^{n}   \mu_{i}\mu{j} \psi^{\dagger}_{t}[i] \psi^{\dagger}_{t}[j]   \right | 1_{\left \{ \delta_i = 0, \delta_j = 0 \right \} } \nonumber \\
    & +  \sigma^{2} \sum_{i,j=1}^{N(JQ+1)}   \left | \sum_{t=1}^{n} \psi^{\dagger}_{t}[i] \psi^{\dagger}_{t}[j]) \psi_i^{T} \psi_j  \right | 1_{\left \{ \delta_i =1,\delta_j =1 \right \}}
\end{align}
Now let's,
\begin{equation}
\mathcal{R}^{U}= \sum_{i,j=1}^{N(JQ+1)}  \left |  \sum_{t=1}^{n}   \mu_{i}\mu{j} \psi^{\dagger}_{t}[i] \psi^{\dagger}_{t}[j]   \right |,
\end{equation}
and,
\begin{equation}
\mathcal{R}^{S}= \sigma^{2} \sum_{i,j=1}^{N(JQ+1)}   \left | \sum_{t=1}^{n} \psi^{\dagger}_{t}[i] \psi^{\dagger}_{t}[j]) \psi_i^{T} \psi_j  \right |.
\end{equation}
Then, based on the following min-max formulation, we obtain an upper bound of the ideal risk, that, when minimized will approximate the ideal risk in the overcomplete case:
\begin{align}
     \mathcal{R}^{\star}(x,W)  & \leq  \sum_{k =1}^{N(JQ+1)} \min_{\delta_k} \max_{\delta_l, l\neq k} \mathcal{R}^{U} 1_{\left \{ \delta_i = 0, \delta_j = 0 \right \} } + \mathcal{R}^{S} 1_{\left \{ \delta_i =1,\delta_j =1 \right \}} \\
     & \leq  \sum_{k =1}^{N(JQ+1)} \min_{\delta_k}  (\max_{\delta_l, l\neq k} \mathcal{R}_1 + \max_{\delta_l, l\neq k}  \mathcal{R}_2) \\
     & =  \sum_{k =1}^{N(JQ+1)} \min_{\delta_k}  \: \: 1_{\left \{ \delta_k =0 \right \}} \left [ \sum_{j =1}^{n*(JQ+1)} \left |  \sum_{t=1}^{n}   \mu_{k}\mu{j} \psi^{\dagger}_{t}[k] \psi^{\dagger}_{t}[j]   \right |\right ] \nonumber \\
     & \: + 1_{\left \{ \delta_k =1 \right \}}  \left [ \sigma^{2} \sum_{j=1}^{N(JQ+1)}   \left | \sum_{t=1}^{n} \psi^{\dagger}_{t}[k] \psi^{\dagger}_{t}[j]) \psi_k^{T} \psi_j  \right | \right ].
\end{align}
Now, let's denote by $\mathcal{R}_{up}^{U}$ the error term corresponding to unselected coefficients:
\begin{equation}
    \mathcal{R}_{up}^{U} =\sum_{j =1}^{n*(JQ+1)} \left | \mu_{k}\mu{j} \sum_{t=1}^{n}    \psi^{\dagger}_{t}[k] \psi^{\dagger}_{t}[j]   \right |,
\end{equation}
and by $\mathcal{R}_{up}^{S}$ for the selected ones:
\begin{equation}
    \mathcal{R}_{up}^{S} = \sigma^{2} \sum_{j=1}^{N(JQ+1)}   \left | \sum_{t=1}^{n} (\psi^{\dagger}_{t}[k] \psi^{\dagger}_{t}[j]) \psi_k^{T} \psi_j  \right |.
\end{equation}
we have that,
\begin{align}
    \mathcal{R}_{up}(x,W) & =  \sum_{k =1}^{N(JQ+1)} \min_{\delta_k}  \: \: 1_{\left \{ \delta_k =0 \right \}} \mathcal{R}_{up}^{U}  + 1_{\left \{ \delta_k =1 \right \}} \mathcal{R}_{up}^{S}  \\
    & =  \sum_{k =1}^{N(JQ+1)}  \min(\mathcal{R}_{up}^{U},\mathcal{R}_{up}^{S}).
\end{align}

\subsection{Comparison Upper Bound Ideal Risk with Orthogonal Ideal Risk}
\label{upper-bound-comparison}
\textbf{Proposition 1.}
\proof
The comparison of this upper bound risk given an orthogonal dictionary and the one derived in the orthogonal case is as follows:

 If the basis is orthogonal, we have,
\begin{equation}
    \sum_{t=1}^{n} (\psi^{\dagger}_{t}[k] \psi^{\dagger}_{t}[j]) = \left\{\begin{matrix}1,\: k=j \\ 
0, \: else 
\end{matrix}\right.
\end{equation}
and,
\begin{equation}
    \psi_k^{T} \psi_j =  \left\{\begin{matrix}1,\: k=j \\ 
0, \: else 
\end{matrix}\right.
\end{equation}
Therefore, the upper-bound derived recovers the ideal risk in the orthogonal case.

\subsection{Comparison Upper Bound Ideal Risk with Empirical Risk}
\label{upper-bound-empirical}
\textbf{Proposition 2.}
\proof

If $D^S=I$, the empirical risk is equal to:
\begin{equation*}
   \tilde{\mathcal{R}}(y,W) = \sigma^{2} \sum_{j=1}^{N(JQ+1)}   \left | \sum_{t=1}^{n} (\psi^{\dagger}_{t}[k] \psi^{\dagger}_{t}[j]) \psi_k^{T} \psi_j  \right |.
\end{equation*}   
and the upper bound risk is:   
\begin{equation*} 
   \mathcal{R}_{up}(x,W) = \sigma^{2} \sum_{j=1}^{N(JQ+1)}   \left | \sum_{t=1}^{n} (\psi^{\dagger}_{t}[k] \psi^{\dagger}_{t}[j]) \psi_k^{T} \psi_j  \right |.
\end{equation*}
Thus both coincide as this restriction on the support of the risk makes it independent of both $x$ and $y$.

\textbf{Proposition 3.}
\proof
In the case where $D^U=I$,
\begin{align}
    \tilde{\mathcal{R}}(y,W) &= \sum_{j=1}^{N(JQ+1)} \left | \mu_k(y) \mu_j(y) \sum_t \psi^{\dagger}_t [k]  \psi^{\dagger}_t [j] \right | \\
    & = \sum_j  \left | ( \mu_k(x) \mu_j(x) +  \mu_k(x) \mu_j(\epsilon) +  \mu_k(\epsilon) \mu_k(x) + \mu_k(\epsilon) \mu_j(\epsilon) ) \right | \times \\
    & \quad \left | \sum_t \psi^{\dagger}_t [k]  \psi^{\dagger}_t [j] \right |,
\end{align}
by the triangular inequality, we have that:
\begin{align}
    \tilde{\mathcal{R}}(y,W)\leq &  \sum_j ( \left | \mu_k(x) \mu_j(x) \right | +\left |  \mu_k(x) \mu_j(\epsilon) \right | +  \left | \mu_k(\epsilon) \mu_k(x)\right | + \left | \mu_k(\epsilon) \mu_j(\epsilon) ) \right | \times \\
    & \qquad \left | \sum_t \psi^{\dagger}_t [k]  \psi^{\dagger}_t [j] \right |,
\end{align}
by the monotony of expectation and the Fubini theorem, we have almost surely:

\begin{equation*}
   \tilde{\mathcal{R}}(y,W)\leq \mathcal{R}_{up}(x,W) + C \times \left | \sum_{t=1}^{n}    \psi^{\dagger}_{t}[k] \psi^{\dagger}_{t}[j]   \right | \qquad a.s.,
\end{equation*}
where $C$ is equals to,
\begin{equation*}
    C =  \sum_{k=1}^{N(JQ+1)} \sum_{j=1}^{N(JQ+1)}   \left | \mu_{k}(x) \right | \left \| \psi_j \right \|_1 \sigma \sqrt{\frac{2}{\pi}} +  \left | \mu_{j}(x) \right |\left \| \psi_k \right \|_1 \sigma \sqrt{\frac{2}{\pi}}  + \sigma^2 (1- \frac{2}{\pi}).
\end{equation*}
\end{document}